\documentclass{ifm-tr}
\usepackage{mathpazo}
\ifPDFTeX\else
  \defaultfontfeatures{Ligatures=TeX}%
  \IfFontExistsTF{Palatino Linotype}{\setmainfont{Palatino Linotype}}{%
    \setmainfont{texgyrepagella}[Extension=.otf, UprightFont=*-regular, BoldFont=*-bold,
      ItalicFont=*-italic, BoldItalicFont=*-bolditalic]}
\fi

\usepackage{amsmath,amssymb,mathtools}
\usepackage{enumitem}
\usepackage{xspace}
\usepackage{url}
\usepackage{fontawesome5}
\newcommand{\titlesubtitlelist}{}
\newcommand{\titlesubtitle}[1]{\def\titlesubtitlelist{#1}}
\renewcommand{\title}[1]{\def\titlelist{#1}}

\newcommand{\patchpartititle}[2]{%
  \patchcmd{\mymaketitle}{#1}{#2}{}{%
    \PackageError{parti-title}{The IFM title template changed}{Check title_layout.tex against ifm-tr.cls.}%
  }%
}
\patchpartititle
  {{\fontsize{19pt}{16pt}\selectfont\sffamily\bfseries\color{black}\titlelist}\par}
  {{\rmfamily\fontsize{21pt}{24pt}\selectfont
      \bfseries\color{black}\titlelist\par}%
   \ifdefempty{\titlesubtitlelist}{}{%
     \vskip 0.10cm
     {\rmfamily\fontsize{12pt}{15pt}\selectfont
       \mdseries\color{ifmmedgray}\titlesubtitlelist\par}%
   }}
\patchpartititle{\vspace{0.15cm}}{\vspace{0.10cm}}
\patchpartititle{\vspace{0.5cm}}{\vspace{0.28cm}}
\patchpartititle{\vskip 0.3cm}{\vskip 0.24cm}
\patchpartititle{\vskip 0.15cm}{\vskip 0.10cm}
\patchpartititle{\vskip 0.05cm}{\vskip 0.04cm}
\patchpartititle{\vskip 0.25cm}{\vskip 0.18cm}
\patchpartititle{\vskip 0.3cm}{\vskip 0.22cm}

\definecolor{sg}{HTML}{8A8A8A}
\usepackage{listings}
\newcommand{\T}[1]{#1}
\newcommand{\F}[1]{#1}
\newcommand{\rms}{\operatorname{rms}}

\newsavebox{\besidetext}
\newsavebox{\besidecap}
\newlength{\besidefig}
\newcommand{\figbeside}[5]{%
  \sbox{\besidetext}{\begin{minipage}[t]{0.60\linewidth}\vspace{0pt}#5\end{minipage}}%
  \sbox{\besidecap}{\begin{minipage}[t]{0.37\linewidth}\vspace{0pt}\captionof{figure}{#3}\end{minipage}}%
  \addtocounter{figure}{-1}%
  \setlength{\besidefig}{\dimexpr\ht\besidetext+\dp\besidetext-\ht\besidecap-\dp\besidecap\relax}%
  \def\besidefigure{\begin{minipage}[t]{0.37\linewidth}\vspace{0pt}\centering
    \includegraphics[width=\linewidth,height=\besidefig,keepaspectratio]{#2}\par
    \captionof{figure}{#3}\label{#4}\end{minipage}}%
  \par\noindent\ifx l#1\besidefigure\hfill\usebox{\besidetext}\else\usebox{\besidetext}\hfill\besidefigure\fi\par}
\definecolor{cin}{HTML}{2E7D32}
\definecolor{cfit}{HTML}{C62828}
\definecolor{cfix}{HTML}{1565C0}
\newcounter{algo}  %
\newcommand{\In}[1]{{\color{cin}#1}}
\newcommand{\Fit}[1]{{\color{cfit}#1}}
\newcommand{\Fix}[1]{{\color{cfix}#1}}
\setlist{itemsep=1pt,topsep=2pt,leftmargin=*}
\title{Latent Feedback Is a Compute and Data Amplifier}
\titlesubtitle{Scaling Law for Latent Feedback Transformers on Repeated Data}
\metadata[\faGithub\ Code]{\url{https://github.com/ifm-ai/LoRM}}
\date{November 11, 2026}
\hypersetup{pdftitle={Latent Feedback Is a Compute and Data Amplifier: Scaling Law for Latent Feedback Transformers on Repeated Data},pdfauthor={Benhao Huang, Chaitanya Dwivedi, Xuezhe Ma, Alexander Moreno, Hector Liu}}
\abstract{
When high-quality data runs out, a common practice is to train on it for several epochs, but each repeat is worth less than the last.
Latent feedback, which feeds each token's top-layer state into the next token's input, is another way to spend that compute, yet it has not been tested against a vanilla Transformer that spends the same compute on more epochs.
We run this test across a ladder of model sizes, unique data and epochs, tuning both models at every size and amount of unique data.
Latent feedback makes every FLOP and every token count for more except when unique data is scarce relative to model size, a regime that today’s models are far from.
We find that latent feedback serves as an amplifier: with unique data fixed, it reaches a Transformer's loss with less compute; with epochs fixed, it reaches that loss with less unique data.
Furthermore, both savings grow with unique data per parameter and with repetition, and downstream tasks agree.
To go beyond our grid, we fit one scaling law to both models, in which latent feedback enters as a Transformer with more parameters; it predicts that both savings keep growing with unique data and repetition. Fitted on models from 60M to 271M parameters, the law predicts the losses at a held-out 778M size to within 0.3\% and the gap between the two models to within 0.05\%. At most, latent feedback saves 48\% of the compute and 47\% of the unique data.
When data, not compute, is the limit, latent feedback is the better use of a FLOP for a model of a given size.

}
\author[\(\ddagger\ast\)]{\href{https://huskydoge.github.io/}{Benhao Huang}}\authorsep{, }
\author[\(\ddagger\)]{Chaitanya Dwivedi}\authorsep{, }
\author[\(\ddagger\dagger\)]{Xuezhe Ma}\authorsep{, }
\author[\(\ddagger\dagger\)]{Alexander Moreno}\authorsep{, }
\author[\(\ddagger\)]{Hector Liu}\authorsep{}
\affiliation[\(\ddagger\)]{Institute of Foundation Models}\affiliationsep{, }
\affiliation[\(\dagger\)]{USC}\affiliationsep{, }
\affiliation[\(\ast\)]{CMU}

\begin{document}
\maketitle
\noindent\begin{minipage}{\linewidth}
\centering
\includegraphics[width=\linewidth]{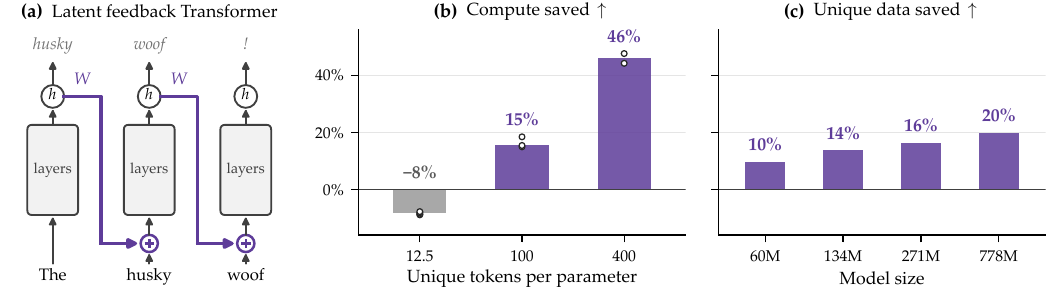}
\vspace{-8pt}
\captionof{figure}{\textbf{Latent feedback saves compute and unique data.}
(a) Latent feedback: each token's top-layer state $h$ enters the next token's input through a map $W$.
(b) The share of compute that latent feedback saves at equal unique data, at a budget of two Transformer epochs; each bar is the median over model sizes, and each dot is one model size.
(c) The share of unique data it saves when both models train one epoch on 50 unique tokens per parameter.}
\label{fig:teaser}
\end{minipage}
\vspace{4pt}
\section{Introduction}\label{sec:intro}
High-quality text, not compute, now limits pretraining of large language models~\citep{chowdhery2022palm,kimi2025k2}.
Even web text, mostly \textbf{not} high quality, grows far more slowly than pretraining compute~\citep{kim2025pretraining,villalobos2024run}.
So a common practice is to train on the best sources for several epochs~\citep{brown2020language,hoffmann2022training,lozhkov2024starcoder2,abdin2024phi4,olmo2025olmo3,microsoftai2026maithinking1}.
However, each repeat is worth less than the last, so additional epochs bring diminishing returns~\citep{lozhkov2024starcoder2,benallal2025smollm2,microsoftai2026maithinking1,muennighoff2023scaling}.

{Latent feedback offers another use for that compute.}
A latent feedback Transformer feeds each token's top-layer state into the next token's input.
Prior work reports improved performance for extra training FLOPs~\citep{fan2020feedback,wang2026fullbandwidth}.
Yet these comparisons fix the training tokens, and none repeats data.
With repeated data, the fair baseline is a Transformer that spends the same FLOPs on more epochs.
Whether latent feedback's gain survives at scale, against a vanilla Transformer given the same FLOPs and unique data, is still largely unknown.
We therefore ask:

\begin{center}
\emph{\underline{When data runs out, can latent feedback make every FLOP and every token count for more?}}
\end{center}

We answer it on a ladder of models from 60M to 778M parameters, each trained on 12.5 to 400 unique tokens per non-embedding parameter (UTPP) for one to four epochs, that is up to 1,600 training tokens per parameter with repetitions (TPP).
Both models are tuned at every size and amount of unique data, and the pools of unique tokens are nested and deduplicated.
The answer is yes, once each parameter has enough unique data, and today's models have far more than enough.
We measure the answer at a fixed model size in two ways, and then fit a law to extrapolate it.
\begin{itemize}
\item \textbf{Compute.} With unique data fixed, the Transformer must train more epochs to match latent feedback; it needs $\kappa$ times the FLOPs (Figure~\ref{fig:compute} (a)).
The saving grows with UTPP and with epochs: at two epochs, latent feedback costs 8\% more at 12.5 UTPP, but saves 15\% to 18\% at 100 and 44\% to 48\% at 400.
Above about 75 UTPP it saves compute at every budget we train, and at two epochs already above 33.
Practice is far past this point: Chinchilla used 20 TPP~\citep{hoffmann2022training}, whereas today the median 1B--10B dense model trains on 1,040, at least 260 UTPP even over four epochs~\citep{grattafiori2024llama3,yang2025qwen3,gemma2025gemma3}.\footnote{Median over the 1B--10B dense models among 257 base models whose reports state both their parameter and training token counts; Appendix~\ref{app:practice} gives the full statistics.}
\item \textbf{Data.} With epochs fixed, the Transformer needs $\mu$ times the unique data to match latent feedback with the same parameters and key-value (KV) cache (Figure~\ref{fig:data} (a)).
At one epoch on 50 UTPP, latent feedback saves 16\% of the unique data at 271M and 20\% at 778M; with three epochs on 100 UTPP, it saves 41\% at 271M.
\item \textbf{A fitted law for extrapolation.} To go beyond our grid, we fit one scaling law to both models, with latent feedback entering as a Transformer with $v$ times the parameters (Section~\ref{sec:law}).
Fitted on 60M to 271M, it predicts the held-out 778M losses to within 0.3\% and the gap between the models to within 0.05\%.
It then draws the boundary between the two regimes, a FLOP spent on another epoch or on latent feedback, and predicts that both savings grow with unique data and repetition beyond the grid.
\end{itemize}

\section{Background}\label{sec:fbt}
Latent Feedback Transformers feed each position's top-layer state into the input of the next position~\citep{fan2020feedback,wang2026fullbandwidth}. At decoding time, position $t$ enters the blocks $\T{f}$ with input $u_t$ and leaves as $h_t$:
\begin{equation}
u_t = e_t + \F{W}\,\rms(h_{t-1}). \label{eq:decode}
\end{equation}
However, teacher-forced training cannot run the positions one by one to get top-layer states, so it runs passes over the whole sequence as fixed-point iterations: 
\begin{equation}
\Big[\,u^{(p+1)}_t = e_t + \F{W}\,\rms\big(h^{(p)}_{t-1}\big),\quad h^{(p)}_{t-1}=\T{f}\big(u^{(p)}_{t-1}\big)\,\Big] \;\xrightarrow{\;p\,\to\,\infty\;}\; u^{(*)}_t = e_t + \F{W}\,\rms\big(h^{(*)}_{t-1}\big), \label{eq:fbt}
\end{equation}

The first pass reads no feedback, $u^{(1)}_t = e_t$, and pass $p = 2, 3, \dots$ reads the previous pass's top states, ideally till fixed points $h^{(*)}_{t-1}=h_{t-1}$. In practice, two to three passes come close to the fixed point~\citep{wang2026fullbandwidth}.

At inference, autoregressive decoding already has the previous token's top state when it predicts the current token, so each generated token costs what a Transformer's does, though prefilling a prompt needs additional passes of Equation~\ref{eq:fbt}. The KV cache is a Transformer's plus one vector.

\noindent
\begin{minipage}[c]{0.55\linewidth}
\textbf{Training.}
Each update runs one pass with probability $p$ and two passes otherwise (Algorithm~\ref{alg:ours}).
The loss and gradient are computed only on the last pass.
The map $W$ is drawn from a normal distribution with standard deviation 0.02.
Because of the additional forward passes, a token with feedback costs $c > 1$ times a Transformer's training FLOPs.
So $p$ trades that cost against the feedback each update sees.
\end{minipage}\hfill
\begin{minipage}[c]{0.4\linewidth}
\refstepcounter{algo}\label{alg:ours}{\small\textbf{Algorithm~\thealgo}\enspace One training update}
\vspace{-4pt}
\begin{lstlisting}[language=python]
x = embed(tokens)
if rand() > p:
    with no_grad():
        s = blocks(x)
    # token t reads s[t-1]
    x = x + W(rms_norm(shift(s)))
loss = ce(head(blocks(x)), y)
loss.backward()
\end{lstlisting}
\end{minipage}%
\hspace{0.025\linewidth}

\section{Scaling law}\label{sec:scaling}
\textbf{Budget.}
A run has $N$ non-embedding parameters~\citep{kaplan2020scaling,li2025steplaw,kimi2025linear}, $U$ unique tokens and a compute budget $F$; models are named by their total parameters.
We count $F$ in Transformer epochs, its FLOPs over those of one Transformer epoch on $U$, so a Transformer trains $R = F$ epochs.
Latent feedback with a training schedule $s$ that costs $c_s > 1$ Transformer FLOPs per token trains only $R = F/c_s$ epochs.
So at equal compute, latent feedback sees fewer repeats than the Transformer.
Appendix~\ref{app:notation} lists the notation.

\textbf{Law.}
We extend the data-constrained law of \citet{muennighoff2023scaling}:
\begin{equation}
L_s(N, U, F) = E + \Big[\Big(\frac{A}{(v_s N)^{\alpha}}\Big)^{q} + \Big(\frac{B}{D_s^{\beta}}\Big)^{q}\Big]^{1/q}, \qquad
D_s = U \Big[1 + R^\ast\big(1 - e^{-(R-1)/R^\ast}\big)\Big], \qquad R = \frac{F}{c_s}. \label{eq:law}
\end{equation}
where $E$ is the irreducible loss, $A/(v_s N)^{\alpha}$ is the capacity term, and $B/D_s^{\beta}$ is the data term.
The factor $v_s$ counts the effective parameters of training schedule $s$, with $v = 1$ for the Transformer.
$D_s$ is the effective number of training tokens, where $R^\ast$ sets how fast repeats lose value, and $q$ couples the two terms.\footnote{Table~\ref{tab:forms} in Appendix~\ref{app:fit} compares three ways to combine the two terms. Simply adding them ($q = 1$) leaves an error about three times larger. Our fit gives $q \approx 3.7$, so the larger term dominates the loss: whichever of model size and data is scarcer limits the run. This form is the law of \citet{videau2026skaling} with $k = 1/q$, written so that $\alpha$ and $\beta$ stay each term's own exponent. Tying $q$ to $1/\beta$, as \citet{kaplan2020scaling} do, leaves an error about six times larger.}

\textbf{Repeats.}
Epoch $k$ is worth about $e^{-(k-1)/R^\ast}$ of a fresh one, so the effective data $D_s$ saturates at $U(1+R^\ast)$. Below one epoch we set $D_s = RU$.
The two pieces join smoothly: near $R = 1$, $1 - e^{-(R-1)/R^\ast} \approx (R-1)/R^\ast$, so both give $D_s \approx RU$ with the same slope.
Following \citet{qin2026bridging}, we let $R^\ast \propto (U/N)^{\rho}$ .

\textbf{Capacity.}
A schedule reruns the same blocks on a token, as a recurrent-depth model reruns its layers~\citep{dehghani2019universal,geiping2025recurrent}.
Scaling laws for such models count the extra passes as effective parameters~\citep{schwethelm2026isodepth,xiong2026truncate}. We follow their practice and introduce $v_s$ that multiplies the parameter count $N$,
\begin{equation}
v_s = v_0 \Big(\frac{D_s}{100\,N}\Big)^{\omega} \Big(\frac{N}{10^{8}}\Big)^{\gamma}, \label{eq:v}
\end{equation}
where $D_s$ is the effective data of Equation~\ref{eq:law} and $v_0$, $\omega$ and $\gamma$ are fitted.\footnote{Appendix~\ref{app:fit} compares this form with a constant $v$ and with a $v$ that grows only with data per parameter; this form fits best and predicts the held-out 778M best (Table~\ref{tab:forms}).}
The constants 100 and $10^8$ serve as a reference point: $v_0$ is the factor of a model with $10^8$ parameters trained on 100 effective tokens per parameter, and $\omega$ and $\gamma$ give how it grows with data per parameter and with size from there.

\textbf{The law predicts a crossover.}
To first order in $\ln v_s$ and $\ln c_s$, a schedule's gap to the Transformer at equal $F$ is (Derivations in Appendix~\ref{app:crossover})
\begin{equation}
L_s - L_T \approx S\,\Big[-\,w\,\alpha\ln v_s \;+\; (1-w)\,\bar\varepsilon\,\beta\ln c_s\Big], \qquad
w = \frac{X^q}{X^q + Y^q}, \qquad \bar\varepsilon = \frac{\ln D_T - \ln D_s}{\ln c_s} \in (0, 1], \label{eq:gap}
\end{equation}
where $L_T$ is the Transformer's loss (Equation~\ref{eq:law} with $v = c = 1$); $X = A/N^{\alpha}$ and $Y = B/D_T^{\beta}$ are its two terms, $S = (X^q + Y^q)^{1/q}$, and $D_T$ and $D_s$ are the effective data of the Transformer and of the schedule.

The first term is the extra capacity, weighted by $w$, which nears 1 when the capacity term dominates.
The second term, weighted by $1 - w$, is the data that latent feedback gives up.
At equal $F$ it trains $c_s$ times fewer epochs, so its effective data $D_s$ is smaller than the Transformer's $D_T$.
How much smaller depends on what the dropped epochs were worth, which $\bar\varepsilon$ measures: $\bar\varepsilon = 1$ when each dropped epoch was worth a fresh one, so $D_s = D_T / c_s$, and $\bar\varepsilon \to 0$ when repeats are worth nothing, so dropping them costs no effective data.
With $U$ fixed, $D$ grows with $F$ (Equation~\ref{eq:law}), so $w$ rises toward 1 and repeats saturate ($\bar\varepsilon \to 0$).
As the budget grows, the capacity term thus counts for more and the data term for less.
So feedback that costs more compute per token, may lose at small budgets but win past a crossover budget $F^\ast$.

\textbf{Two multipliers} express latent feedback's advantage as the compute or the unique data it saves.
Both compare the two models at the same size $N$ and the same validation loss.
The \emph{compute multiplier} $\kappa$ is the Transformer's budget over latent feedback's, at equal unique data $U$, so latent feedback saves a share $1 - 1/\kappa$ of the compute.
The \emph{data multiplier} $\mu$ is the Transformer's unique data over latent feedback's, at equal epochs $R$, so latent feedback saves a share $1 - 1/\mu$ of the data:
\begin{equation}
\kappa = \frac{F_T}{F} \ \text{ with } L_T(N, U, F_T) = L_s(N, U, F), \qquad
\mu = \frac{U_T}{U} \ \text{ with } L_T(N, U_T, R) = L_s(N, U, c_s R). \label{eq:multipliers}
\end{equation}

\textbf{Fitting.}
We fit all cells of both models jointly by least squares.
Latent feedback differs from the Transformer only through its capacity factor $v_s$, with three fitted coefficients ($v_0$, $\omega$ and $\gamma$), and its measured cost $c_s$; all other coefficients are shared.\footnote{A two-phase fit, which first fits $E, A, \alpha, B, \beta$ on one-epoch Transformer runs~\citep{muennighoff2023scaling,lovelace2026prescriptive}, fits worse than our joint fit in both the additive and the coupled form (Table~\ref{tab:forms} in Appendix~\ref{app:fit}).} To separate the effect of the size and data terms, our cells vary training tokens per parameter from 12.5 to 1,600~\citep{kricheli2026identifiability}.

\section{Setup}\label{sec:setup}
\textbf{Models.}
Both models, the Transformer and the one with latent feedback, share one backbone at 60M, 134M, 271M and 778M parameters.
The backbone is a modern decoder: pre-norm RMSNorm~\citep{zhang2019rmsnorm}, rotary positions~\citep{su2021roformer}, QK-norm~\citep{henry2020qknorm} and SwiGLU~\citep{shazeer2020glu}, with untied embeddings and OLMo~2's initialization~\citep{olmo2025olmo2}.
Its matrices train with NorMuon~\citep{li2025normuon}, a normalized variant of Muon~\citep{jordan2024muon}, and its embeddings and output head with AdamW~\citep{loshchilov2019adamw}; see Table~\ref{tab:config} in Appendix~\ref{app:config} for details.  Latent feedback uses $p = 0.15$ throughout\footnote{We find that its loss at equal compute barely moves with $p$ when around $0.15$ (Appendix~\ref{app:config}).}, so its cost $c$ is about 1.20 to 1.27 across the four sizes.

\textbf{Data.}
We take the Dolma~3 sample~\citep{olmo2025olmo3} and process it ourselves: we deduplicate it and remove the documents that share a paragraph with Paloma~\citep{magnusson2023paloma}.
This keeps Paloma a clean held-out benchmark for a fair evaluation~\citep{soldaini2024dolma,groeneveld2024olmo}.
We order its documents so that every prefix has the whole sample's mixture of sources and topics with same ratios, and each run's pool is the first $U$ tokens of this order (Appendix~\ref{app:pools}).
Pools are thus nested and differ only in size, from $U = 12.5N$ to $400N$.

\textbf{Tuning.}
We call one setting of model size $N$, unique data $U$, and budget $F$ a cell.
For each model, size and amount of unique data, we tune a main run over the learning rate $\eta$ and the timescale $\tau = 1/(\eta\lambda T)$~\citep{wang2024adamw,bergsma2025powerlines}, where $\lambda$ is the weight-decay and $T$ the number of steps; other $F$ branch off the main run as decay runs and keep its hyperparameters.
Each grid is centered on a predicted start point from earlier runs and extended wherever the best point sits on its edge; a cell's value is the minimum of a quadratic fitted to its points.
Where unique data is small, we also shrink the batch to stay near its optimum~\citep{bergsma2025powerlines}. Each main run uses a warmup-stable-decay schedule~\citep{hagele2024scaling} (Appendix~\ref{app:config}).

\textbf{Evaluation.}
We measure on a held-out validation set of 47,274 documents (100M tokens), taken from the front of the same stratified order of the deduplicated Dolma~3 sample. It keeps the mixture ratio and excludes training documents. For downstream tasks, we score the Olmo~3 Base Easy suite~\citep{olmo2025olmo3}, 109 tasks in 37 suites, as bits per byte (BPB) of the correct continuation on OLMES requests~\citep{gu2024olmes}.
Each suite is weighted by the inverse of its seed variance, so noisy suites do not dominate (Appendix~\ref{app:checks}).

Decoding latent feedback token by token (Equation~\ref{eq:decode}) is slow.
We score it instead with 4 parallel passes of Equation~\ref{eq:fbt}, whose states $h^{(4)}$ nearly match the decoding states, so more passes leave the loss almost unchanged (Table~\ref{tab:passes} in Appendix~\ref{app:passes}).

\section{Experiment and result}\label{sec:results}
This section reads the grid of runs from Section~\ref{sec:setup} on validation loss in three steps, and then checks it on downstream tasks.
First, with unique data fixed, we measure the compute that latent feedback saves ($\kappa$, Section~\ref{sec:flops}).
Second, with the epochs fixed, we measure the unique data that it saves ($\mu$, Section~\ref{sec:data}).
Third, we fit the law of Section~\ref{sec:scaling} (Equation~\ref{eq:law}) jointly to both models, test it on the held-out 778M, and use it to explain the two savings and to predict them beyond the grid (Section~\ref{sec:law}).
Finally, we check that the comparison is robust: repeats do not overfit, gaps are stable across seeds, and downstream tasks agree with the validation loss (Section~\ref{sec:downstream}).

\subsection{Equal unique data: the compute multiplier}\label{sec:flops}
\begin{figure}[tb]
\centering
\includegraphics[width=\linewidth]{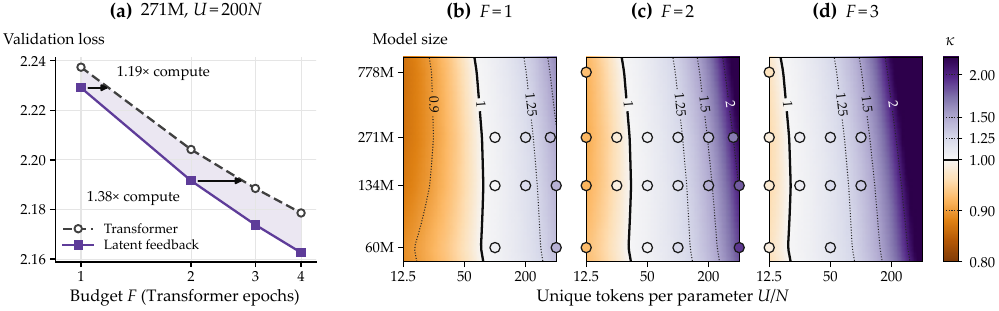}
\vspace{-8pt}
\caption{\textbf{The compute multiplier $\kappa$ across unique data and model size.}
(a) 271M with $U = 200N$: each arrow runs from a latent feedback cell to the budget at which the Transformer reaches its loss, labelled with $\kappa$.
(b)--(d) Dots are measured cells, filled with the $\kappa$ read off the Transformer's losses at $F = 1$ to 4 interpolated in $\log F$; everywhere else the map shows the law of Section~\ref{sec:law}. $\kappa$ grows with unique tokens per parameter and with the budget $F$, and its contours run nearly vertically: it depends on $U/N$ far more than on size.}
\label{fig:compute}
\end{figure}
\textbf{At equal size and unique data, latent feedback reaches a Transformer's loss with less compute} (Figure~\ref{fig:compute} (a)).
Both models train the same model size $N$ on the same $U$ unique tokens, and only the budget $F$ varies: the Transformer trains $F$ epochs and latent feedback, about 1.2 times as costly per token, trains $F/c_s$.
Latent feedback saves a share $1 - 1/\kappa$ of the compute (Equation~\ref{eq:multipliers}).

\textbf{The saving grows with UTPP and with epochs} (Figure~\ref{fig:compute}).
Along UTPP, latent feedback costs more compute than the Transformer when unique data is scarce and saves compute when it is plentiful.
The unique data at which both models need the same compute ($\kappa = 1$) falls as the budget grows: about 75 UTPP at $F = 1$ and 33 at $F = 2$.
Above $\approx$75 UTPP, latent feedback saves compute at every budget we train.
Along epochs, the saving grows with the budget $F$.
At 400 UTPP and $F = 2$, latent feedback saves nearly half the compute ($\kappa = 1.91$ at 60M); Table~\ref{tab:kappa} in Appendix~\ref{app:cells} lists every measured $\kappa$.

\subsection{Equal epochs: the data multiplier}\label{sec:data}
\begin{figure}[tb]
\centering
\includegraphics[width=\linewidth]{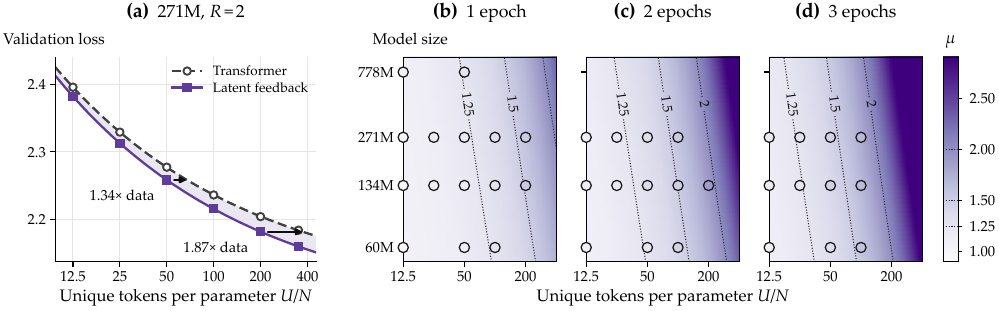}
\vspace{-8pt}
\caption{\textbf{The data multiplier $\mu$ across unique data and model size.}
(a) 271M at two epochs: each arrow is the unique data latent feedback saves; the arrow at $U = 200N$ is extrapolated. (b)--(d) Dots are measured cells, read off the Transformer's losses across $U$ interpolated in $\log U$ (Appendix~\ref{app:cells}); everywhere else, the map shows the law of Section~\ref{sec:law}.
$\mu$ grows with unique data and with epochs, and depends on $U/N$ far more than on size.}
\label{fig:data}
\end{figure}
\textbf{At equal size and epochs, latent feedback reaches a Transformer's loss with up to half the unique data} (Figure~\ref{fig:data} (a)).
With size and epochs equal, both serve with the same parameters and KV cache; latent feedback pays only about 1.2 times the training FLOPs per token, and saves $1 - 1/\mu$ of the unique data. At 271M and two epochs, for example, $\mu = 1.34$ at $U = 50N$, so latent feedback needs 25\% less unique data than the vanilla Transformer (Figure~\ref{fig:data} (a)).

\textbf{The saving grows with $U/N$ and with epochs} (Figure~\ref{fig:data}).
At the smallest $U$ the two models need about the same data.
The largest measured saving at 271M is 41\% of the unique data, with $U = 100N$ and three epochs ($\mu = 1.69$).
At one epoch it also grows with size. Table~\ref{tab:mu} in Appendix~\ref{app:cells} lists every measured $\mu$.

\subsection{One law for both models}\label{sec:law}
\begin{table}[tb]
\centering\footnotesize\setlength{\tabcolsep}{6pt}
\caption{\textbf{Held-out test of the law at 778M.}
Fitted on 60M to 271M only. Error: (predicted $-$ measured) / measured.}
\label{tab:holdout}
\begin{tabular}{llcccccc}
\toprule
 & & \multicolumn{3}{c}{Transformer} & \multicolumn{3}{c}{Latent feedback} \\
\cmidrule(lr){3-5}\cmidrule(lr){6-8}
$U$ & $F$ & Measured & Predicted & Error (\%) & Measured & Predicted & Error (\%) \\
\midrule
12.5$N$ & 1 & 2.2054 & 2.2112 & $+$0.26 & 2.2211 & 2.2276 & $+$0.29 \\
12.5$N$ & 2 & 2.1434 & 2.1460 & $+$0.12 & 2.1501 & 2.1532 & $+$0.15 \\
12.5$N$ & 3 & 2.1136 & 2.1172 & $+$0.17 & 2.1162 & 2.1195 & $+$0.16 \\
12.5$N$ & 4 & 2.0950 & 2.1005 & $+$0.26 & 2.0950 & 2.0996 & $+$0.22 \\
50$N$ & 1 & 2.0878 & 2.0873 & $-$0.03 & 2.0893 & 2.0897 & $+$0.02 \\
50$N$ & 2 & 2.0452 & 2.0454 & $+$0.01 & 2.0421 & 2.0413 & $-$0.04 \\
100$N$ & 1 & 2.0448 & 2.0441 & $-$0.03 & 2.0413 & 2.0406 & $-$0.03 \\
\bottomrule
\end{tabular}%

\end{table}
The two multipliers above are measured only inside the grid, up to 778M, 400 UTPP and four epochs.
To explain them and to go beyond the grid, we fit Equation~\ref{eq:law} jointly to both models' losses at 60M, 134M and 271M, which gives
\begin{gather}
L = 0.5 + \Big[\Big(\frac{1.75}{(v\,n)^{0.112}}\Big)^{3.75} + \Big(\frac{1.40}{d^{\,0.126}}\Big)^{3.75}\Big]^{1/3.75}, \qquad
R^\ast = 14.6\,\Big(\frac{u}{n}\Big)^{0.45}, \label{eq:fitted}\\
v = 1 \ \text{ for the Transformer}, \qquad v = 1.14\,\Big(\frac{d}{n}\Big)^{0.010} n^{0.007} \ \text{ for latent feedback}, \notag
\end{gather}
with $n = N/10^{8}$, $u = U/10^{10}$ and $d = D/10^{10}$, where the effective data $D$ follows Equation~\ref{eq:law}.\footnote{Table~\ref{tab:symbols} in Appendix~\ref{app:notation} lists every coefficient and its meaning and fitted value.}

\textbf{The law predicts a held-out size} (Table~\ref{tab:holdout}).
The 778M model is 3.3 times the size of the largest fitted one, yet the law predicts its losses with a mean absolute error of 0.13\% and at most 0.29\%.
This error is below those that recent scaling laws report on their own held-out runs, 0.50\% for Farseer and 0.60\% for Skaling~\citep{li2025farseer,videau2026skaling}.
The law also picks the winner: it is off by about the same amount for both models, so it predicts the gap to within 0.05\% in every held-out row, and gives the measured sign in 6 of 7, with the seventh a tie, and in all but two of the fitted cells, both at $U = 12.5N$ and $F = 4$, where the predicted gap is within 0.001 of zero.

With the law checked at 778M, we use it to explain the two savings, to draw the boundary between the two regimes, and to predict the savings beyond the grid.
\begin{figure}[tb]
\centering
\includegraphics[width=\linewidth]{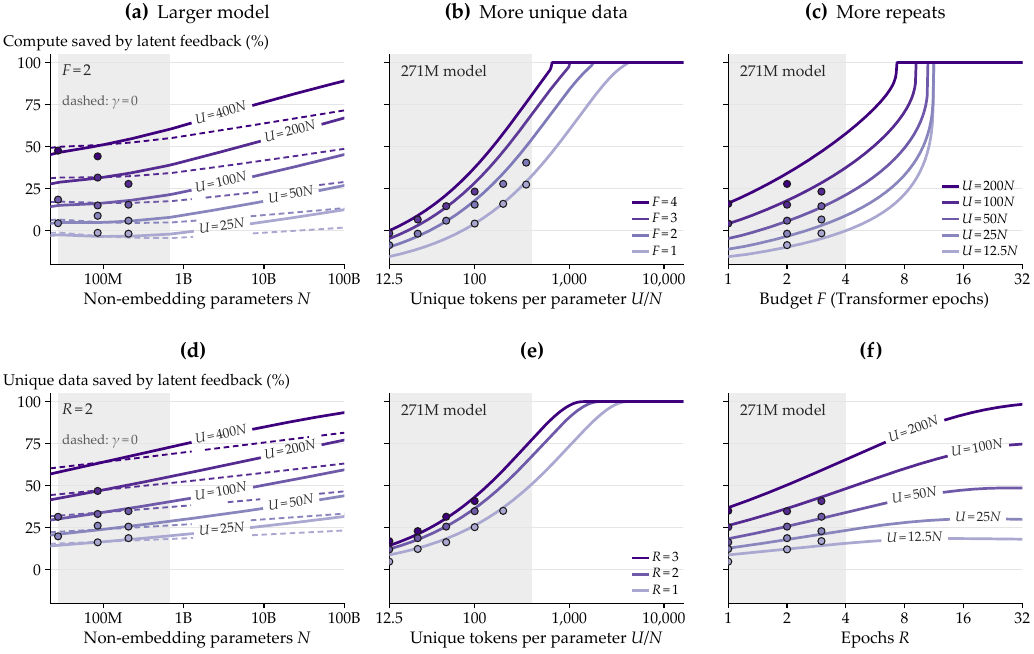}
\vspace{-8pt}
\caption{\textbf{The savings beyond the grid, as the law of Section~\ref{sec:law} predicts them.} Every curve comes from one fit of the law to all cells; the dots are the few of those cells that lie on each curve. Each panel extends one variable beyond its measured range (shaded), with one curve per value of a second variable; in (a) and (d), dashed curves set $\gamma = 0$, so $v$ does not grow with size. Top: compute saved at equal unique data. Bottom: unique data saved at equal epochs. A saving of 100\% means the Transformer cannot reach latent feedback's loss with any number of epochs or any amount of unique data.}
\label{fig:extrap}
\end{figure}

\textbf{First, the law explains the two savings: latent feedback trades epochs for capacity.}
Equation~\ref{eq:fitted} has a capacity term and a data term, and latent feedback moves both.
It lowers the capacity term, because it fits as a Transformer with 1.14 times the parameters, a factor that grows slowly with data per parameter and with size.
However, it raises the data term, because its cost leaves it about 1.2 times fewer epochs, and repeated epochs still count: in the coupled fits, four epochs at 134M are worth 3.0 to 3.9 fresh ones.
Latent feedback therefore wins where capacity limits the loss, and loses where data limits it.
This is why both savings grow with unique data and with more epochs: each moves the Transformer toward its capacity floor, which latent feedback lowers.

We want to point out the savings hold at a fixed model size.
Latent feedback spends $c$ times a Transformer's training FLOPs per token, as much as a Transformer $c$ times larger, yet the law counts it as a Transformer only $v$ times larger, and $v$ stays below $c$ (1.10 to 1.19 against 1.20 to 1.27).
So when model size is free, simply spending the same FLOPs on a Transformer $c$ times larger, with the same unique data, gives a loss 0.006 to 0.012 lower in every fitted cell.
That Transformer, however, costs $c$ times more per decoding step and holds a larger KV cache, while latent feedback serves at the cost of the smaller model.

\textbf{Second, the law draws the boundary between the two regimes} (the line $\kappa = 1$ in Figure~\ref{fig:compute}).\footnote{Figure~\ref{fig:decision} in Appendix~\ref{app:fit} maps the loss gap behind it.}
On one side of this line, a FLOP is worth more as another Transformer epoch; on the other, it is worth more as latent feedback.
The boundary moves toward less unique data as the budget grows: latent feedback wins from about 75 UTPP at $F = 1$, 33 at $F = 2$, 20 at $F = 3$ and 12 at $F = 4$.

\textbf{Third, the law predicts that both savings keep growing beyond the grid} (Figure~\ref{fig:extrap}; Table~\ref{tab:extrap} in Appendix~\ref{app:fit}).
Each panel extends one variable beyond the measured range.
Along size, the savings rise because $\alpha < \beta$ makes larger models more capacity-limited, and further through $\gamma$: at 100B, 100 UTPP and $F = 2$, the law predicts $\kappa = 1.83$, and 1.41 with $\gamma = 0$.
Within 60M to 271M, $\gamma$ is not significantly above zero ($p = 0.054$), but the held-out 778M favors it: without $\gamma$, the law underrates latent feedback in all seven held-out rows.
The dashed curves in Figure~\ref{fig:extrap} show the savings if $v$ does not grow with size.
More unique data per parameter raises them fast, until the Transformer cannot catch up at any budget: it nears its capacity floor, while latent feedback's floor is that of a Transformer about 1.2 times larger.
More repeats push the compute saving the same way, since repeats saturate for both models and only the capacity gap remains; the data saving instead levels off at a height set by the unique data.
The curves only shift with the other variables, so latent feedback passes the crossover at any unique data once a run repeats enough, as long as repeats do not overfit (true up to four epochs; Figure~\ref{fig:noover} in Appendix~\ref{app:cells}).

\subsection{Robustness: repeats, seeds and downstream tasks}\label{sec:downstream}
\begin{table}[tb]
\centering\footnotesize\setlength{\tabcolsep}{5pt}
\caption{\textbf{Downstream tasks at equal model size, unique data and compute.}
Bits per byte (BPB $\downarrow$) of both models at budgets of $F = 1$ to 4 Transformer epochs.
The lower of each pair is shaded.
BPB averages 37 task suites, each weighted by its inverse seed variance (Appendix~\ref{app:checks}).  }
\label{tab:main}
\setlength{\fboxsep}{1.5pt}%
\begin{tabular}{llcccccccc}
\toprule
 & & \multicolumn{2}{c}{$F = 1$} & \multicolumn{2}{c}{$F = 2$} & \multicolumn{2}{c}{$F = 3$} & \multicolumn{2}{c}{$F = 4$} \\
\cmidrule(lr){3-4}\cmidrule(lr){5-6}\cmidrule(lr){7-8}\cmidrule(lr){9-10}
Size & $U$ & Transformer & Feedback & Transformer & Feedback & Transformer & Feedback & Transformer & Feedback \\
\midrule
60M & 12.5$N$ & \colorbox{black!9}{1.2324} & 1.2453 & \colorbox{black!9}{1.1623} & 1.1680 & 1.1399 & \colorbox[HTML]{D9D0E8}{1.1388} & 1.1279 & \colorbox[HTML]{D9D0E8}{1.1186} \\
 & 50$N$ & 1.1051 & \colorbox[HTML]{D9D0E8}{1.1014} & 1.0664 & \colorbox[HTML]{D9D0E8}{1.0539} & 1.0511 & \colorbox[HTML]{D9D0E8}{1.0371} & 1.0413 & \colorbox[HTML]{D9D0E8}{1.0278} \\
 & 100$N$ & 1.0673 & \colorbox[HTML]{D9D0E8}{1.0566} & 1.0373 & \colorbox[HTML]{D9D0E8}{1.0205} & 1.0220 & \colorbox[HTML]{D9D0E8}{1.0072} & 1.0163 & \colorbox[HTML]{D9D0E8}{0.9999} \\
 & 400$N$ & 1.0116 & \colorbox[HTML]{D9D0E8}{0.9917} & 0.9970 & \colorbox[HTML]{D9D0E8}{0.9708} & 0.9867 & \colorbox[HTML]{D9D0E8}{0.9634} & 0.9822 & \colorbox[HTML]{D9D0E8}{0.9615} \\
\midrule
134M & 12.5$N$ & \colorbox{black!9}{1.0374} & 1.0514 & 0.9907 & \colorbox[HTML]{D9D0E8}{0.9889} & 0.9677 & \colorbox[HTML]{D9D0E8}{0.9624} & 0.9559 & \colorbox[HTML]{D9D0E8}{0.9466} \\
 & 25$N$ & 0.9865 & \colorbox[HTML]{D9D0E8}{0.9835} & 0.9428 & \colorbox[HTML]{D9D0E8}{0.9423} & 0.9263 & \colorbox[HTML]{D9D0E8}{0.9225} & 0.9151 & \colorbox[HTML]{D9D0E8}{0.9088} \\
 & 50$N$ & 0.9444 & \colorbox[HTML]{D9D0E8}{0.9380} & 0.9160 & \colorbox[HTML]{D9D0E8}{0.9062} & 0.9013 & \colorbox[HTML]{D9D0E8}{0.8880} & 0.8930 & \colorbox[HTML]{D9D0E8}{0.8766} \\
 & 100$N$ & 0.9071 & \colorbox[HTML]{D9D0E8}{0.9020} & 0.8849 & \colorbox[HTML]{D9D0E8}{0.8767} & 0.8739 & \colorbox[HTML]{D9D0E8}{0.8602} & 0.8648 & \colorbox[HTML]{D9D0E8}{0.8544} \\
 & 200$N$ & 0.8838 & \colorbox[HTML]{D9D0E8}{0.8725} & 0.8637 & \colorbox[HTML]{D9D0E8}{0.8484} & 0.8527 & \colorbox[HTML]{D9D0E8}{0.8389} & 0.8501 & \colorbox[HTML]{D9D0E8}{0.8309} \\
 & 400$N$ & 0.8634 & \colorbox[HTML]{D9D0E8}{0.8541} & 0.8506 & \colorbox[HTML]{D9D0E8}{0.8348} & 0.8412 & \colorbox[HTML]{D9D0E8}{0.8282} & 0.8370 & \colorbox[HTML]{D9D0E8}{0.8213} \\
\midrule
271M & 12.5$N$ & \colorbox{black!9}{0.9096} & 0.9173 & \colorbox{black!9}{0.8665} & 0.8675 & 0.8508 & \colorbox[HTML]{D9D0E8}{0.8470} & 0.8380 & \colorbox[HTML]{D9D0E8}{0.8337} \\
 & 25$N$ & \colorbox{black!9}{0.8678} & 0.8690 & 0.8325 & \colorbox[HTML]{D9D0E8}{0.8282} & 0.8179 & \colorbox[HTML]{D9D0E8}{0.8119} & 0.8073 & \colorbox[HTML]{D9D0E8}{0.8012} \\
 & 50$N$ & 0.8336 & \colorbox[HTML]{D9D0E8}{0.8309} & 0.8045 & \colorbox[HTML]{D9D0E8}{0.8009} & 0.7902 & \colorbox[HTML]{D9D0E8}{0.7854} & 0.7844 & \colorbox[HTML]{D9D0E8}{0.7729} \\
 & 100$N$ & 0.8021 & \colorbox[HTML]{D9D0E8}{0.7980} & 0.7792 & \colorbox[HTML]{D9D0E8}{0.7723} & 0.7675 & \colorbox[HTML]{D9D0E8}{0.7582} & 0.7619 & \colorbox[HTML]{D9D0E8}{0.7502} \\
 & 200$N$ & 0.7783 & \colorbox[HTML]{D9D0E8}{0.7712} & 0.7603 & \colorbox[HTML]{D9D0E8}{0.7499} & 0.7499 & \colorbox[HTML]{D9D0E8}{0.7392} & 0.7430 & \colorbox[HTML]{D9D0E8}{0.7308} \\
 & 350$N$ & 0.7652 & \colorbox[HTML]{D9D0E8}{0.7525} & 0.7469 & \colorbox[HTML]{D9D0E8}{0.7365} & 0.7394 & \colorbox[HTML]{D9D0E8}{0.7265} & 0.7375 & \colorbox[HTML]{D9D0E8}{0.7217} \\
\midrule
778M & 12.5$N$ & \colorbox{black!9}{0.7670} & 0.7715 & \colorbox{black!9}{0.7301} & 0.7322 & 0.7134 & \colorbox[HTML]{D9D0E8}{0.7106} & 0.7009 & \colorbox[HTML]{D9D0E8}{0.6985} \\
 & 50$N$ & 0.6959 & \colorbox[HTML]{D9D0E8}{0.6949} & 0.6673 & \colorbox[HTML]{D9D0E8}{0.6662} & --- & --- & --- & --- \\
 & 100$N$ & 0.6683 & \colorbox[HTML]{D9D0E8}{0.6613} & --- & --- & --- & --- & --- & --- \\
\bottomrule
\end{tabular}%

\end{table}
\textbf{Repeats up to four epochs do not overfit either model} (Figure~\ref{fig:noover} in Appendix~\ref{app:cells}).
Even at the scarcest unique data, validation loss keeps falling through four epochs at every size.
The gaps are stable across seeds: at 134M, a gap's standard deviation is below 0.002 and none changes sign (Appendix~\ref{app:cells}).

\textbf{Downstream tasks agree well with the validation loss} (Table~\ref{tab:main}).
In Table~\ref{tab:main}, the Transformer has the lower BPB only at the smallest unique data ($U \le 25N$) and budgets of $F \le 2$; latent feedback has it everywhere else, so the winner switches at almost the same place as in Figure~\ref{fig:compute}.
At $F = 4$, latent feedback has the lower BPB in all 17 finished cells.
On downstream tasks, latent feedback starts to win slightly earlier than on validation loss, and the two metrics pick the same winner in 59 of 71 paired cells (Table~\ref{tab:cells} in Appendix~\ref{app:cells} gives the validation losses).

\section{Related work}\label{sec:related}
\textbf{Spending compute on scarce data.}
With unique data scarce, prior work fits a scaling law and, given unique data, compute, and a fixed architecture, chooses the model size and number of epochs~\citep{muennighoff2023scaling,qin2026bridging,lovelace2026prescriptive}. When taking architecture into account, sparsity in models delays the point where repeats saturate~\citep{wu2026sparse};
masked diffusion~\citep{prabhudesai2025diffusion} overtakes autoregressive training once compute is abundant, across unique data and epochs.
However, its gain needs excessive epochs: on 100M unique tokens its best loss comes at 500 epochs, against 50 for autoregressive training, whose hyperparameters follow a prior recipe instead of being tuned per run.
The looped model Parcae~\citep{prairie2026parcae} spends extra FLOPs on loops at a fixed parameter count, yet weighs them against fresh tokens rather than epochs.
Latent feedback pays off within four epochs once each parameter has enough unique data, and we price it against epochs on fixed unique data, tuning both models at every size and amount of unique data.

\textbf{Recurrence in depth.}
Looped models buy more compute per token by running their layers several times within each token~\citep{dehghani2019universal,geiping2025recurrent,huang2026topology}, and their scaling laws count the extra loops as effective parameters~\citep{schwethelm2026isodepth,xiong2026truncate,prairie2026parcae}.
Our law takes the same view: latent feedback enters it as a Transformer with $v$ times the parameters.
However, a looped model still pays its extra compute again at every decoding step, even if it allows KV cache sharing~\citep{huang2026towards}.
Latent feedback recurs across tokens instead: it pays for its extra passes in training and in prefill, and generates each token like a Transformer with one more cached vector.

\textbf{Feedback across tokens.}
Feedback memory~\citep{fan2020feedback} first fed deep states to later tokens but trained token by token; recent models feed back a deeper layer~\citep{wang2026fullbandwidth,huang2026lrt,cai2026t2mlr,mozer2026recirculation} or a frozen teacher's state~\citep{tirosh2026lift}. At equal compute or training time, some beat a Transformer~\citep{wang2026fullbandwidth,huang2026lrt} and others trail it~\citep{cai2026t2mlr,tirosh2026lift}.
Yet none varies data repetition: even \citet{wang2026fullbandwidth}, who motivate feedback by scarce unique data, give the Transformer more tokens rather than more epochs.
We compare against more epochs, across model size, unique data and repetition, and fit all three in one scaling law.

\section{Conclusion}
We show that latent feedback makes every FLOP and every token count for more: once each parameter has enough unique data, it reaches a given loss with less compute than more epochs, and with less unique data at equal epochs.
Both savings grow with unique data and repetition, and one scaling law, fitted to latent feedback and the Transformer together, predicts them at a held-out size.
As a takeaway, when high-quality data, not compute, limits a run of a given model size, feeding back the top-layer state uses compute better than training more epochs.

\paragraph{Limitation.}
So far, our ladder stops at a relatively small scale and uses a single backbone; larger models and data budgets, as well as other architectures may shift the crossover. 
We also train latent feedback with a single fixed schedule, and further optimizations are possible. Lastly, a comparison with the recurrent depth models will be of great value. We leave these to future work.

\paragraph{Acknowledgment.}
We sincerely thank Zitian Gao and Xi Wang for many valuable discussions.

\FloatBarrier
\begingroup
\small
\setlength{\bibsep}{0pt}
\bibliographystyle{unsrtnat}
\bibliography{proposal}

\begin{thebibliography}{70}
\providecommand{\natexlab}[1]{#1}
\providecommand{\url}[1]{\texttt{#1}}
\expandafter\ifx\csname urlstyle\endcsname\relax
  \providecommand{\doi}[1]{doi: #1}\else
  \providecommand{\doi}{doi: \begingroup \urlstyle{rm}\Url}\fi

\bibitem[Chowdhery et~al.(2022)Chowdhery, Narang, Devlin, Bosma, Mishra, Roberts, Barham, Chung, Sutton, Gehrmann, et~al.]{chowdhery2022palm}
Aakanksha Chowdhery, Sharan Narang, Jacob Devlin, Maarten Bosma, Gaurav Mishra, Adam Roberts, Paul Barham, Hyung~Won Chung, Charles Sutton, Sebastian Gehrmann, et~al.
\newblock {PaLM}: Scaling language modeling with pathways.
\newblock arXiv:2204.02311, 2022.

\bibitem[{Kimi Team} et~al.(2025){Kimi Team}, Bai, Bao, Charles, Chen, Chen, Chen, Chen, Chen, Chen, et~al.]{kimi2025k2}
{Kimi Team}, Yifan Bai, Yiping Bao, Y.~Charles, Cheng Chen, Guanduo Chen, Haiting Chen, Huarong Chen, Jiahao Chen, Ningxin Chen, et~al.
\newblock Kimi {K2}: Open agentic intelligence.
\newblock arXiv:2507.20534, 2025.

\bibitem[Kim et~al.(2025)Kim, Kotha, Liang, and Hashimoto]{kim2025pretraining}
Konwoo Kim, Suhas Kotha, Percy Liang, and Tatsunori Hashimoto.
\newblock Pre-training under infinite compute.
\newblock arXiv:2509.14786, 2025.

\bibitem[Villalobos et~al.(2024)Villalobos, Ho, Sevilla, Besiroglu, Heim, and Hobbhahn]{villalobos2024run}
Pablo Villalobos, Anson Ho, Jaime Sevilla, Tamay Besiroglu, Lennart Heim, and Marius Hobbhahn.
\newblock Will we run out of data? limits of {LLM} scaling based on human-generated data.
\newblock arXiv:2211.04325, 2024.
\newblock Version 2.

\bibitem[Brown et~al.(2020)Brown, Mann, Ryder, Subbiah, Kaplan, Dhariwal, Neelakantan, Shyam, Sastry, Askell, et~al.]{brown2020language}
Tom~B. Brown, Benjamin Mann, Nick Ryder, Melanie Subbiah, Jared Kaplan, Prafulla Dhariwal, Arvind Neelakantan, Pranav Shyam, Girish Sastry, Amanda Askell, et~al.
\newblock Language models are few-shot learners.
\newblock arXiv:2005.14165, 2020.

\bibitem[Hoffmann et~al.(2022)Hoffmann, Borgeaud, Mensch, Buchatskaya, Cai, Rutherford, Casas, Hendricks, Welbl, Clark, et~al.]{hoffmann2022training}
Jordan Hoffmann, Sebastian Borgeaud, Arthur Mensch, Elena Buchatskaya, Trevor Cai, Eliza Rutherford, Diego de~Las Casas, Lisa~Anne Hendricks, Johannes Welbl, Aidan Clark, et~al.
\newblock Training compute-optimal large language models.
\newblock \emph{arXiv preprint arXiv:2203.15556}, 2022.

\bibitem[Lozhkov et~al.(2024)Lozhkov, Li, Ben~Allal, Cassano, Lamy-Poirier, Tazi, Tang, Pykhtar, Liu, Wei, et~al.]{lozhkov2024starcoder2}
Anton Lozhkov, Raymond Li, Loubna Ben~Allal, Federico Cassano, Joel Lamy-Poirier, Nouamane Tazi, Ao~Tang, Dmytro Pykhtar, Jiawei Liu, Yuxiang Wei, et~al.
\newblock {StarCoder} 2 and the stack v2: The next generation.
\newblock arXiv:2402.19173, 2024.

\bibitem[Abdin et~al.(2024{\natexlab{a}})Abdin, Aneja, Behl, Bubeck, Eldan, Gunasekar, Harrison, Hewett, Javaheripi, Kauffmann, et~al.]{abdin2024phi4}
Marah Abdin, Jyoti Aneja, Harkirat Behl, S{\'e}bastien Bubeck, Ronen Eldan, Suriya Gunasekar, Michael Harrison, Russell~J. Hewett, Mojan Javaheripi, Piero Kauffmann, et~al.
\newblock Phi-4 technical report.
\newblock arXiv:2412.08905, 2024{\natexlab{a}}.

\bibitem[{Team Olmo} et~al.(2025){Team Olmo}, Ettinger, Bertsch, Kuehl, Graham, Heineman, Groeneveld, Brahman, Timbers, Ivison, et~al.]{olmo2025olmo3}
{Team Olmo}, Allyson Ettinger, Amanda Bertsch, Bailey Kuehl, David Graham, David Heineman, Dirk Groeneveld, Faeze Brahman, Finbarr Timbers, Hamish Ivison, et~al.
\newblock {Olmo} 3.
\newblock arXiv:2512.13961, 2025.

\bibitem[{The Microsoft AI Team}(2026)]{microsoftai2026maithinking1}
{The Microsoft AI Team}.
\newblock {MAI-Thinking-1}: Building a hill-climbing machine.
\newblock \url{https://microsoft.ai/pdf/mai-thinking-1.pdf}, 2026.
\newblock Technical report.

\bibitem[Ben~Allal et~al.(2025)Ben~Allal, Lozhkov, Bakouch, Mart{\'i}n~Bl{\'a}zquez, Penedo, Tunstall, Marafioti, Kydl{\'i}{\v{c}}ek, Piqueres~Lajar{\'i}n, Srivastav, et~al.]{benallal2025smollm2}
Loubna Ben~Allal, Anton Lozhkov, Elie Bakouch, Gabriel Mart{\'i}n~Bl{\'a}zquez, Guilherme Penedo, Lewis Tunstall, Andr{\'e}s Marafioti, Hynek Kydl{\'i}{\v{c}}ek, Agust{\'i}n Piqueres~Lajar{\'i}n, Vaibhav Srivastav, et~al.
\newblock {SmolLM2}: When smol goes big -- data-centric training of a small language model.
\newblock arXiv:2502.02737, 2025.

\bibitem[Muennighoff et~al.(2023)Muennighoff, Rush, Barak, Scao, Piktus, Tazi, Pyysalo, Wolf, and Raffel]{muennighoff2023scaling}
Niklas Muennighoff, Alexander~M. Rush, Boaz Barak, Teven~Le Scao, Aleksandra Piktus, Nouamane Tazi, Sampo Pyysalo, Thomas Wolf, and Colin Raffel.
\newblock Scaling data-constrained language models.
\newblock arXiv:2305.16264, 2023.

\bibitem[Fan et~al.(2020)Fan, Lavril, Grave, Joulin, and Sukhbaatar]{fan2020feedback}
Angela Fan, Thibaut Lavril, Edouard Grave, Armand Joulin, and Sainbayar Sukhbaatar.
\newblock Addressing some limitations of transformers with feedback memory.
\newblock arXiv:2002.09402, 2020.

\bibitem[Wang et~al.(2026)Wang, Cai, Zhan, Dong, Fan, de~Rosa, Pearce, and Langford]{wang2026fullbandwidth}
Xi~Wang, Ziyang Cai, Zheng Zhan, Harry Dong, Ying Fan, Gustavo de~Rosa, Tim Pearce, and John Langford.
\newblock Full-bandwidth transformer.
\newblock arXiv:2608.08888, 2026.

\bibitem[Grattafiori et~al.(2024)Grattafiori, Dubey, Jauhri, Pandey, Kadian, et~al.]{grattafiori2024llama3}
Aaron Grattafiori, Abhimanyu Dubey, Abhinav Jauhri, Abhinav Pandey, Abhishek Kadian, et~al.
\newblock The {Llama} 3 herd of models.
\newblock arXiv:2407.21783, 2024.

\bibitem[Yang et~al.(2025)Yang, Li, Yang, Zhang, Hui, et~al.]{yang2025qwen3}
An~Yang, Anfeng Li, Baosong Yang, Beichen Zhang, Binyuan Hui, et~al.
\newblock {Qwen3} technical report.
\newblock arXiv:2505.09388, 2025.

\bibitem[{Gemma Team} et~al.(2025){Gemma Team}, Kamath, Ferret, Pathak, Vieillard, et~al.]{gemma2025gemma3}
{Gemma Team}, Aishwarya Kamath, Johan Ferret, Shreya Pathak, Nino Vieillard, et~al.
\newblock Gemma 3 technical report.
\newblock arXiv:2503.19786, 2025.

\bibitem[Kaplan et~al.(2020)Kaplan, McCandlish, Henighan, Brown, Chess, Child, Gray, Radford, Wu, and Amodei]{kaplan2020scaling}
Jared Kaplan, Sam McCandlish, Tom Henighan, Tom~B. Brown, Benjamin Chess, Rewon Child, Scott Gray, Alec Radford, Jeffrey Wu, and Dario Amodei.
\newblock Scaling laws for neural language models.
\newblock arXiv:2001.08361, 2020.

\bibitem[Li et~al.(2025{\natexlab{a}})]{li2025steplaw}
Houyi Li et~al.
\newblock Predictable scale: Part {I}, {Step Law} -- optimal hyperparameter scaling law in large language model pretraining.
\newblock arXiv:2503.04715, 2025{\natexlab{a}}.

\bibitem[{Kimi Team}(2025)]{kimi2025linear}
{Kimi Team}.
\newblock {Kimi Linear}: An expressive, efficient attention architecture.
\newblock arXiv:2510.26692, 2025.

\bibitem[Videau et~al.(2026)Videau, Youbi-Idrissi, Lopez-Paz, and Ahuja]{videau2026skaling}
Mathurin Videau, Badr Youbi-Idrissi, David Lopez-Paz, and Kartik Ahuja.
\newblock Skaling: {Chinchilla}'s exponents meet {Kaplan}'s coupling.
\newblock arXiv:2608.07222, 2026.

\bibitem[Qin et~al.(2026)Qin, Hamidieh, and Alvarez-Melis]{qin2026bridging}
Tian Qin, Kimia Hamidieh, and David Alvarez-Melis.
\newblock Bridging compute- and data-optimal pretraining.
\newblock arXiv:2607.25271, 2026.

\bibitem[Dehghani et~al.(2019)Dehghani, Gouws, Vinyals, Uszkoreit, and Kaiser]{dehghani2019universal}
Mostafa Dehghani, Stephan Gouws, Oriol Vinyals, Jakob Uszkoreit, and {\L}ukasz Kaiser.
\newblock Universal transformers.
\newblock In \emph{International Conference on Learning Representations}, 2019.

\bibitem[Geiping et~al.(2025)]{geiping2025recurrent}
Jonas Geiping et~al.
\newblock Scaling up test-time compute with latent reasoning: A recurrent depth approach.
\newblock arXiv:2502.05171, 2025.

\bibitem[Schwethelm et~al.(2026)Schwethelm, R{\"u}ckert, et~al.]{schwethelm2026isodepth}
Kristian Schwethelm, Daniel R{\"u}ckert, et~al.
\newblock How much is one recurrence worth? iso-depth scaling laws for looped language models.
\newblock arXiv:2604.21106, 2026.

\bibitem[Xiong et~al.(2026)Xiong, Chen, Shao, Su, Jin, Luo, Yang, and Liu]{xiong2026truncate}
Xuyuan Xiong, Siyu Chen, Zihan Shao, Yue Su, Weiyang Jin, Ping Luo, Zhuoran Yang, and Xihui Liu.
\newblock Truncate or not? a scaling study of looped transformers with truncated backpropagation.
\newblock arXiv:2610.26030, 2026.

\bibitem[Lovelace et~al.(2026)Lovelace, Belardi, Kundurthy, Sudhakar, and Weinberger]{lovelace2026prescriptive}
Justin Lovelace, Christian Belardi, Srivatsa Kundurthy, Shriya Sudhakar, and Kilian~Q. Weinberger.
\newblock Prescriptive scaling laws for data constrained training.
\newblock arXiv:2605.01640, 2026.

\bibitem[Kricheli et~al.(2026)Kricheli, Reid, Sarkar, Gandikota, and Shakarian]{kricheli2026identifiability}
Joshua~Shay Kricheli, Alexander~Lawrence Reid, Soumajyoti Sarkar, Venkata Gandikota, and Paulo Shakarian.
\newblock Tokens-per-parameter coverage is critical for robust {LLM} scaling law extrapolation.
\newblock arXiv:2605.08541, 2026.

\bibitem[Zhang and Sennrich(2019)]{zhang2019rmsnorm}
Biao Zhang and Rico Sennrich.
\newblock Root mean square layer normalization.
\newblock In \emph{Advances in Neural Information Processing Systems}, 2019.

\bibitem[Su et~al.(2021)Su, Lu, Pan, Murtadha, Wen, and Liu]{su2021roformer}
Jianlin Su, Yu~Lu, Shengfeng Pan, Ahmed Murtadha, Bo~Wen, and Yunfeng Liu.
\newblock {RoFormer}: Enhanced transformer with rotary position embedding.
\newblock arXiv:2104.09864, 2021.

\bibitem[Henry et~al.(2020)Henry, Dachapally, Pawar, and Chen]{henry2020qknorm}
Alex Henry, Prudhvi~Raj Dachapally, Shubham~Shantaram Pawar, and Yuxuan Chen.
\newblock Query-key normalization for transformers.
\newblock In \emph{Findings of the Association for Computational Linguistics: EMNLP 2020}, 2020.

\bibitem[Shazeer(2020)]{shazeer2020glu}
Noam Shazeer.
\newblock {GLU} variants improve transformer.
\newblock arXiv:2002.05202, 2020.

\bibitem[{Team OLMo} et~al.(2025){Team OLMo}, Walsh, Soldaini, Groeneveld, Lo, Arora, Bhagia, Gu, Huang, Jordan, et~al.]{olmo2025olmo2}
{Team OLMo}, Pete Walsh, Luca Soldaini, Dirk Groeneveld, Kyle Lo, Shane Arora, Akshita Bhagia, Yuling Gu, Shengyi Huang, Matt Jordan, et~al.
\newblock 2 {OLMo} 2 furious.
\newblock \emph{arXiv preprint arXiv:2501.00656}, 2025.

\bibitem[Li et~al.(2025{\natexlab{b}})Li, Liu, Liang, Chen, and Zhao]{li2025normuon}
Zichong Li, Liming Liu, Chen Liang, Weizhu Chen, and Tuo Zhao.
\newblock {NorMuon}: Making {Muon} more efficient and scalable.
\newblock arXiv:2510.05491, 2025{\natexlab{b}}.

\bibitem[Jordan et~al.(2024)Jordan, Jin, Boza, You, Cesista, Newhouse, and Bernstein]{jordan2024muon}
Keller Jordan, Yuchen Jin, Vlado Boza, Jiacheng You, Franz Cesista, Laker Newhouse, and Jeremy Bernstein.
\newblock {Muon}: An optimizer for hidden layers in neural networks.
\newblock \url{https://kellerjordan.github.io/posts/muon/}, 2024.

\bibitem[Loshchilov and Hutter(2019)]{loshchilov2019adamw}
Ilya Loshchilov and Frank Hutter.
\newblock Decoupled weight decay regularization.
\newblock In \emph{International Conference on Learning Representations}, 2019.

\bibitem[Magnusson et~al.(2023)Magnusson, Bhagia, Hofmann, Soldaini, Jha, Tafjord, Schwenk, Walsh, Elazar, Lo, Groeneveld, Beltagy, Hajishirzi, Smith, Richardson, and Dodge]{magnusson2023paloma}
Ian Magnusson, Akshita Bhagia, Valentin Hofmann, Luca Soldaini, Ananya~Harsh Jha, Oyvind Tafjord, Dustin Schwenk, Evan~Pete Walsh, Yanai Elazar, Kyle Lo, Dirk Groeneveld, Iz~Beltagy, Hannaneh Hajishirzi, Noah~A. Smith, Kyle Richardson, and Jesse Dodge.
\newblock Paloma: A benchmark for evaluating language model fit.
\newblock arXiv:2312.10523, 2023.

\bibitem[Soldaini et~al.(2024)Soldaini, Kinney, Bhagia, et~al.]{soldaini2024dolma}
Luca Soldaini, Rodney Kinney, Akshita Bhagia, et~al.
\newblock Dolma: an open corpus of three trillion tokens for language model pretraining research.
\newblock arXiv:2402.00159, 2024.

\bibitem[Groeneveld et~al.(2024)Groeneveld, Beltagy, Walsh, et~al.]{groeneveld2024olmo}
Dirk Groeneveld, Iz~Beltagy, Pete Walsh, et~al.
\newblock {OLMo}: Accelerating the science of language models.
\newblock arXiv:2402.00838, 2024.

\bibitem[Wang and Aitchison(2024)]{wang2024adamw}
Xi~Wang and Laurence Aitchison.
\newblock How to set {AdamW}'s weight decay as you scale model and dataset size.
\newblock arXiv:2405.13698, 2024.

\bibitem[Bergsma et~al.(2025)Bergsma, Dey, Gosal, Gray, Soboleva, and Hestness]{bergsma2025powerlines}
Shane Bergsma, Nolan Dey, Gurpreet Gosal, Gavia Gray, Daria Soboleva, and Joel Hestness.
\newblock Power lines: Scaling laws for weight decay and batch size in {LLM} pre-training.
\newblock arXiv:2505.13738, 2025.

\bibitem[H{\"a}gele et~al.(2024)H{\"a}gele, Bakouch, Kosson, Ben~Allal, Von~Werra, and Jaggi]{hagele2024scaling}
Alexander H{\"a}gele, Elie Bakouch, Atli Kosson, Loubna Ben~Allal, Leandro Von~Werra, and Martin Jaggi.
\newblock Scaling laws and compute-optimal training beyond fixed training durations.
\newblock arXiv:2405.18392, 2024.

\bibitem[Gu et~al.(2024)Gu, Tafjord, Kuehl, Haddad, Dodge, and Hajishirzi]{gu2024olmes}
Yuling Gu, Oyvind Tafjord, Bailey Kuehl, Dany Haddad, Jesse Dodge, and Hannaneh Hajishirzi.
\newblock {OLMES}: A standard for language model evaluations.
\newblock arXiv:2406.08446, 2024.

\bibitem[Li et~al.(2025{\natexlab{c}})]{li2025farseer}
Houyi Li et~al.
\newblock Predictable scale: Part {II}, {Farseer}: A refined scaling law in large language models.
\newblock arXiv:2506.10972, 2025{\natexlab{c}}.

\bibitem[Wu et~al.(2026)Wu, Xiao, Okanovic, Sternal, van Keulen, Pechenizkiy, Mocanu, Hoefler, and Mocanu]{wu2026sparse}
Boqian Wu, Qiao Xiao, Patrik Okanovic, Tomasz Sternal, Maurice van Keulen, Mykola Pechenizkiy, Elena Mocanu, Torsten Hoefler, and Decebal~Constantin Mocanu.
\newblock When data is scarce: Scaling sparse language models with repeated training.
\newblock In \emph{International Conference on Machine Learning}, 2026.
\newblock arXiv:2606.01155.

\bibitem[Prabhudesai et~al.(2025)Prabhudesai, Wu, Zadeh, Fragkiadaki, and Pathak]{prabhudesai2025diffusion}
Mihir Prabhudesai, Mengning Wu, Amir Zadeh, Katerina Fragkiadaki, and Deepak Pathak.
\newblock Diffusion beats autoregressive in data-constrained settings.
\newblock arXiv:2507.15857, 2025.

\bibitem[Prairie et~al.(2026)Prairie, Novack, Berg-Kirkpatrick, and Fu]{prairie2026parcae}
Hayden Prairie, Zachary Novack, Taylor Berg-Kirkpatrick, and Daniel~Y. Fu.
\newblock Parcae: Scaling laws for stable looped language models.
\newblock arXiv:2604.12946, 2026.

\bibitem[Huang et~al.(2026{\natexlab{a}})Huang, Shi, Chen, Wen, Liu, Xing, and Ma]{huang2026topology}
Benhao Huang, Chufan Shi, Junlin Chen, Shicheng Wen, Zhengzhong Liu, Eric Xing, and Xuezhe Ma.
\newblock Towards looped models done right, part i: Topology, input injection, recurrent-state design.
\newblock \emph{arXiv preprint arXiv:2610.23036}, 2026{\natexlab{a}}.

\bibitem[Huang et~al.(2026{\natexlab{b}})Huang, Shi, Chen, Wen, Liu, Xing, and Ma]{huang2026towards}
Benhao Huang, Chufan Shi, Junlin Chen, Shicheng Wen, Zhengzhong Liu, Eric Xing, and Xuezhe Ma.
\newblock Towards looped models done right, part ii: Rethinking at fixed points.
\newblock \emph{arXiv preprint arXiv:2610.06833}, 2026{\natexlab{b}}.

\bibitem[Huang et~al.(2026{\natexlab{c}})Huang, He, Ren, Wang, Peng, Cheng, Wang, He, Gao, Lee, and Shen]{huang2026lrt}
Zeyi Huang, Xuehai He, LiLiang Ren, Yiping Wang, Baolin Peng, Hao Cheng, Shuohang Wang, Pengcheng He, Jianfeng Gao, Yong~Jae Lee, and Yelong Shen.
\newblock Latent recurrent transformer: Architecture exploration, training strategies, and scaling behavior.
\newblock arXiv:2605.26797, 2026{\natexlab{c}}.

\bibitem[Cai et~al.(2026)Cai, Zhu, Dong, He, and Arora]{cai2026t2mlr}
Ziyang Cai, Xingyu Zhu, Yihe Dong, Yinghui He, and Sanjeev Arora.
\newblock {T$^2$MLR}: Transformer with temporal middle-layer recurrence.
\newblock arXiv:2607.15178, 2026.

\bibitem[Mozer et~al.(2026)Mozer, Siddiqui, Sawyer, Sanyal, and Liu]{mozer2026recirculation}
Michael~C Mozer, Shoaib~Ahmed Siddiqui, Danny Sawyer, Sunny Sanyal, and Rosanne Liu.
\newblock Recirculation.
\newblock \emph{arXiv preprint arXiv:2608.17981}, 2026.

\bibitem[Tirosh et~al.(2026)Tirosh, Amos, and Geva]{tirosh2026lift}
Dor Tirosh, Ido Amos, and Mor Geva.
\newblock Pretraining latent information feedback transformers with teacher supervision.
\newblock arXiv:2609.38149, 2026.

\bibitem[{Qwen Team} et~al.(2024){Qwen Team}, Yang, Yang, Zhang, et~al.]{qwen2024qwen25}
{Qwen Team}, An~Yang, Baosong Yang, Beichen Zhang, et~al.
\newblock {Qwen2.5} technical report.
\newblock arXiv:2412.15115, 2024.

\bibitem[{Meta}(2024{\natexlab{a}})]{meta2024llama32card}
{Meta}.
\newblock {Llama} 3.2 model card.
\newblock \url{https://github.com/meta-llama/llama-models/blob/main/models/llama3_2/MODEL_CARD.md}, 2024{\natexlab{a}}.

\bibitem[Chen et~al.(2026)Chen, Li, Zhou, Gong, Jiang, et~al.]{chen2026minimaxm2}
Aili Chen, Aonian Li, Baichuan Zhou, Bangwei Gong, Binyang Jiang, et~al.
\newblock The {MiniMax-M2} series: Mini activations unleashing max real-world intelligence.
\newblock arXiv:2605.26494, 2026.

\bibitem[Zhang et~al.(2024)Zhang, Zeng, Wang, and Lu]{zhang2024tinyllama}
Peiyuan Zhang, Guangtao Zeng, Tianduo Wang, and Wei Lu.
\newblock {TinyLlama}: An open-source small language model.
\newblock arXiv:2401.02385, 2024.

\bibitem[{Meta}(2024{\natexlab{b}})]{meta2024llama3card}
{Meta}.
\newblock {Llama} 3 model card.
\newblock \url{https://github.com/meta-llama/llama-models/blob/main/models/llama3/MODEL_CARD.md}, 2024{\natexlab{b}}.

\bibitem[{NVIDIA} et~al.(2025){NVIDIA}, Blakeman, Basant, Khattar, et~al.]{nvidia2025nemotronh}
{NVIDIA}, Aaron Blakeman, Aarti Basant, Abhinav Khattar, et~al.
\newblock {Nemotron-H}: A family of accurate and efficient hybrid {Mamba-Transformer} models.
\newblock arXiv:2504.03624, 2025.

\bibitem[{MiniCPM Team} et~al.(2025){MiniCPM Team}, Xiao, Li, Han, Bai, et~al.]{minicpm2025minicpm4}
{MiniCPM Team}, Chaojun Xiao, Yuxuan Li, Xu~Han, Yuzhuo Bai, et~al.
\newblock {MiniCPM4}: Ultra-efficient {LLMs} on end devices.
\newblock arXiv:2506.07900, 2025.

\bibitem[Abdin et~al.(2024{\natexlab{b}})Abdin, Aneja, Awadalla, Awadallah, Awan, et~al.]{abdin2024phi3}
Marah Abdin, Jyoti Aneja, Hany Awadalla, Ahmed Awadallah, Ammar~Ahmad Awan, et~al.
\newblock Phi-3 technical report: A highly capable language model locally on your phone.
\newblock arXiv:2404.14219, 2024{\natexlab{b}}.

\bibitem[{Gemma Team} et~al.(2024){Gemma Team}, Riviere, Pathak, Sessa, Hardin, et~al.]{gemma2024gemma2}
{Gemma Team}, Morgane Riviere, Shreya Pathak, Pier~Giuseppe Sessa, Cassidy Hardin, et~al.
\newblock Gemma 2: Improving open language models at a practical size.
\newblock arXiv:2408.00118, 2024.

\bibitem[{GLM-4.5 Team} et~al.(2025){GLM-4.5 Team}, Zeng, Lv, Zheng, et~al.]{glm2025glm45}
{GLM-4.5 Team}, Aohan Zeng, Xin Lv, Qinkai Zheng, et~al.
\newblock {GLM-4.5}: Agentic, reasoning, and coding ({ARC}) foundation models.
\newblock arXiv:2508.06471, 2025.

\bibitem[{DeepSeek-AI} et~al.(2024{\natexlab{a}}){DeepSeek-AI}, Liu, Feng, Xue, Wang, et~al.]{deepseek2024v3}
{DeepSeek-AI}, Aixin Liu, Bei Feng, Bing Xue, Bingxuan Wang, et~al.
\newblock {DeepSeek-V3} technical report.
\newblock arXiv:2412.19437, 2024{\natexlab{a}}.

\bibitem[Touvron et~al.(2023)Touvron, Martin, Stone, Albert, Almahairi, et~al.]{touvron2023llama2}
Hugo Touvron, Louis Martin, Kevin Stone, Peter Albert, Amjad Almahairi, et~al.
\newblock Llama 2: Open foundation and fine-tuned chat models.
\newblock arXiv:2307.09288, 2023.

\bibitem[Biderman et~al.(2023)Biderman, Schoelkopf, Anthony, Bradley, O'Brien, et~al.]{biderman2023pythia}
Stella Biderman, Hailey Schoelkopf, Quentin Anthony, Herbie Bradley, Kyle O'Brien, et~al.
\newblock Pythia: A suite for analyzing large language models across training and scaling.
\newblock arXiv:2304.01373, 2023.

\bibitem[{DeepSeek-AI} et~al.(2024{\natexlab{b}}){DeepSeek-AI}, Bi, Chen, Chen, et~al.]{deepseek2024llm}
{DeepSeek-AI}, Xiao Bi, Deli Chen, Guanting Chen, et~al.
\newblock {DeepSeek LLM}: Scaling open-source language models with longtermism.
\newblock arXiv:2401.02954, 2024{\natexlab{b}}.

\bibitem[Almazrouei et~al.(2023)Almazrouei, Alobeidli, Alshamsi, Cappelli, Cojocaru, et~al.]{almazrouei2023falcon}
Ebtesam Almazrouei, Hamza Alobeidli, Abdulaziz Alshamsi, Alessandro Cappelli, Ruxandra Cojocaru, et~al.
\newblock The {Falcon} series of open language models.
\newblock arXiv:2311.16867, 2023.

\bibitem[Marek et~al.(2025)Marek, Lotfi, Somasundaram, Wilson, and Goldblum]{marek2025small}
Martin Marek, Sanae Lotfi, Aditya Somasundaram, Andrew~Gordon Wilson, and Micah Goldblum.
\newblock Small batch size training for language models: When vanilla {SGD} works, and why gradient accumulation is wasteful.
\newblock arXiv:2507.07101, 2025.

\bibitem[Pearce and Song(2024)]{pearce2024reconciling}
Tim Pearce and Jinyeop Song.
\newblock Reconciling {Kaplan} and {Chinchilla} scaling laws.
\newblock arXiv:2406.12907, 2024.

\end{thebibliography}
\endgroup

\clearpage
\appendix
\section{Notation}\label{app:notation}
Table~\ref{tab:notation} lists the quantities of the main text, and Table~\ref{tab:symbols} the symbols of the fitted law.

\begin{table}[!ht]
\centering\footnotesize\setlength{\tabcolsep}{6pt}
\caption{Quantities of the main text, in the order they are defined.}
\label{tab:notation}
\begin{tabular}{@{}lll@{}}
\toprule
Quantity & Meaning & Defined in \\
\midrule
TPP & Training tokens per parameter, repeats included & Section~\ref{sec:intro} \\
UTPP & Unique tokens per non-embedding parameter, $U/N$ & Section~\ref{sec:intro} \\
\midrule
Latent feedback Transformer & Algorithm~\ref{alg:ours}: one or two passes, no gradient in the first & Section~\ref{sec:fbt} \\
$p$ & Probability that an update runs one pass & Section~\ref{sec:fbt} \\
\midrule
$F$ & Compute budget, in Transformer epochs & Section~\ref{sec:scaling} \\
$c_s$ & Training FLOPs per token of schedule $s$ over a Transformer's & Section~\ref{sec:scaling} \\
$R$ & Epochs trained within the budget, $F/c_s$ & Section~\ref{sec:scaling} \\
$F^\ast$ & Crossover budget, above which latent feedback wins & Section~\ref{sec:scaling} \\
$\kappa$ & Compute multiplier, at equal size and unique data & Section~\ref{sec:scaling} \\
$\mu$ & Data multiplier, at equal size and epochs & Section~\ref{sec:scaling} \\
\bottomrule
\end{tabular}
\end{table}

\begin{table}[!ht]
\centering\footnotesize\setlength{\tabcolsep}{4pt}
\caption{Symbols of Equation~\ref{eq:law-colored}, with the joint fit on the 128 cells.
Appendix~\ref{app:fit} gives the equation.}
\label{tab:symbols}
\begin{tabular}{@{}lll@{}}
\toprule
Symbol & Meaning & Value or source \\
\midrule
\multicolumn{3}{@{}l}{\In{\textit{Inputs}}} \\
\midrule
\In{$s$} & Training schedule & Transformer, or latent feedback at $p = 0.15$ \\
\In{$N$} & Non-embedding parameters & 27.3M, 85.0M, 205.6M (60M, 134M, 271M) \\
\In{$U$} & Unique tokens & $12.5N$ to $400N$ \\
\In{$F$} & Compute budget, in Transformer epochs & 1 to 4 \\
\midrule
\multicolumn{3}{@{}l}{\Fit{\textit{Fitted coefficients}}} \\
\midrule
\Fit{$E$} & Irreducible loss & 0.5 (its lower bound) \\
\Fit{$A$} & Scale of the parameter term & 1.755 \\
\Fit{$\alpha$} & Exponent of the parameter term & 0.112 \\
\Fit{$B$} & Scale of the data term & 1.401 \\
\Fit{$\beta$} & Exponent of the data term & 0.126 \\
\Fit{$q$} & Coupling of the two terms; $q = 1$ is additive & 3.748 \\
\Fit{$R^\ast_T$} & $R^\ast$ at $U = 100N$ & 14.60 \\
\Fit{$\rho$} & Growth of $R^\ast$ with $U/N$ & $+$0.450 \\
\Fit{$v_0$} & Capacity factor $v_s$ at the reference run & 1.139 \\
\Fit{$\omega$} & Growth of $v_s$ with effective tokens per parameter & 0.010 \\
\Fit{$\gamma$} & Growth of $v_s$ with model size & 0.007 \\
\midrule
\multicolumn{3}{@{}l}{\Fix{\textit{Fixed constants}}} \\
\midrule
\Fix{$c_s$} & Training FLOPs per token over a Transformer's & 1.197, 1.232, 1.251 (60M, 134M, 271M) \\
\Fix{$N_0$, $D_0$} & Reference run: parameters and effective tokens & $10^{8}$ and $10^{10}$ \\
\midrule
\multicolumn{3}{@{}l}{\textit{Derived}} \\
\midrule
$R$ & Epochs within the budget $F$ & $F/c_s$ \\
$n$, $u$, $d_s$ & $N$, $U$ and $D_s$ in reference units & $N/N_0$, $U/D_0$, $D_s/D_0$ \\
$D_s$ & Effective data, in fresh tokens & Equation~\ref{eq:law-colored} \\
$L_s$ & Predicted validation loss at the tuned optimum & Equation~\ref{eq:law-colored} \\
\bottomrule
\end{tabular}%

\end{table}
\FloatBarrier

\section{Repeated data in practice}\label{app:practice}

\begin{figure}[!ht]
\centering
\includegraphics[width=\linewidth]{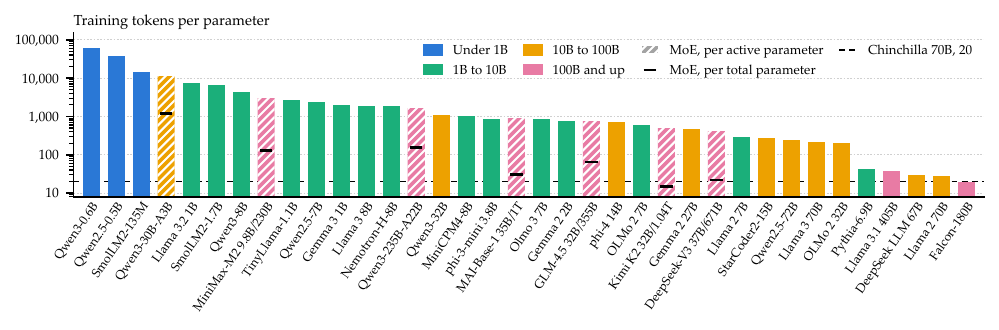}
\vspace{-14pt}
\caption{Training tokens per parameter of base models.
Sources are each developer's report or model card~\citep{yang2025qwen3,qwen2024qwen25,benallal2025smollm2,meta2024llama32card,chen2026minimaxm2,zhang2024tinyllama,gemma2025gemma3,meta2024llama3card,nvidia2025nemotronh,minicpm2025minicpm4,abdin2024phi3,microsoftai2026maithinking1,olmo2025olmo3,gemma2024gemma2,glm2025glm45,abdin2024phi4,olmo2025olmo2,kimi2025k2,deepseek2024v3,touvron2023llama2,lozhkov2024starcoder2,biderman2023pythia,grattafiori2024llama3,deepseek2024llm,almazrouei2023falcon}.
Colour gives the total size, and hatched bars are MoE models.
MoE bars count active parameters; the tick marks total parameters.}
\label{fig:tpp}
\end{figure}

\textbf{Models train far past the compute-optimal ratio} (Figure~\ref{fig:tpp}).
Chinchilla trained on 20 tokens per parameter~\citep{hoffmann2022training}.
We checked 257 base models whose reports state both counts, each against its source.
The median dense model trains on about 6,800 tokens per parameter below 1B.
It uses 1,040 from 1B to 10B, and 210 from 10B to 100B.
The median mixture-of-experts (MoE) model uses 1,640 per active parameter.

\FloatBarrier

\section{Parallel passes}\label{app:passes}
Each pass feeds position $t$ only the states of earlier positions, so after $p$ passes the first $p$ positions equal decoding, and the fixed point of Equation~\ref{eq:fbt} is unique and equals decoding.
More parallel passes than the 4 we evaluate with barely move the loss, far less than the gaps between the two models (Table~\ref{tab:passes}).

\begin{table}[!h]
\centering
\caption{Loss of latent feedback scored with $P$ parallel passes of Equation~\ref{eq:fbt}, on ten runs of an earlier ladder (one 271M, nine 778M) and a 10M-token Dolma~3 validation slice. Differences are taken per run against 16 passes.}
\label{tab:passes}
\begin{tabular}{lccc}
\toprule
Passes $P$ & Mean loss & Mean difference $\downarrow$ & Largest difference $\downarrow$ \\
\midrule
4 & 2.02863 & 0.00007 & 0.00013 \\
8 & 2.02869 & 0.00001 & 0.00004 \\
16 & 2.02870 & -- & -- \\
\bottomrule
\end{tabular}
\end{table}

\section{Every cell}\label{app:cells}
\begin{figure}[!t]
\centering
\includegraphics[width=\linewidth]{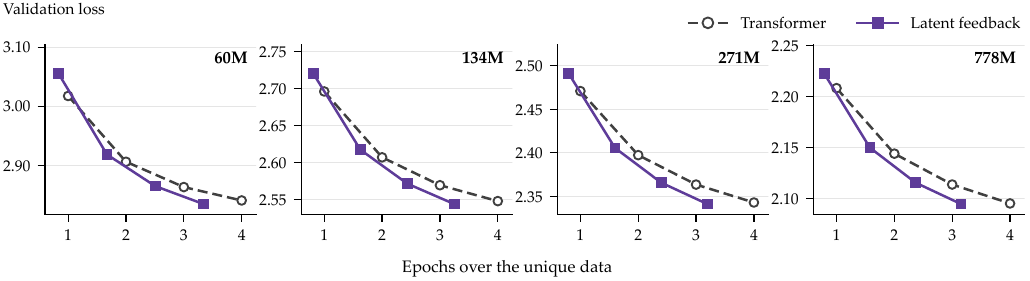}
\caption{\textbf{Repeats up to four epochs do not overfit.}
Validation loss after one to four epochs at the scarcest unique data, $U = 12.5N$, one panel per model size; both curves keep falling.
Latent feedback costs $c > 1$ Transformer FLOPs per token, so at equal compute it trains fewer epochs and its points sit to the left.}
\label{fig:noover}
\end{figure}
Table~\ref{tab:cells} lists the validation loss of every cell at $F = 1$ to 4, and Table~\ref{tab:main} its downstream BPB.

\textbf{Seeds.}
At 134M we ran three seeds at $U = 12.5N$ and $100N$, for $F = 1$ to 4.
The standard deviation of a cell's loss is 0.0003 to 0.0014.
That of a gap is 0.0004 to 0.0017, and every gap keeps its sign.

\textbf{Multipliers.}
Table~\ref{tab:kappa} (top) lists the measured $\kappa$.
Each latent feedback cell's loss is read off the Transformer's losses at the same size and $U$, interpolated in $\log F$ between $F = 1$ and 4.
Table~\ref{tab:mu} (bottom) lists the measured $\mu$.
Latent feedback's loss at $R = 1$, 2, 3 is interpolated in $\log R$ between its branches at $R = F/c$.
We then read $\mu$ off the Transformer's losses across $U$, interpolated in $\log U$ between levels.

\begin{table}[!ht]
\centering\footnotesize\setlength{\tabcolsep}{5pt}
\caption{\textbf{Validation loss $\downarrow$ of every cell at $F = 1$ to 4.}
Both models, at equal size, unique data and budget $F$; Feedback is the latent feedback Transformer.
The lower of each pair is shaded: purple where latent feedback leads, grey where the Transformer does.
778M is held out of the law; ---: its grid has no such cell.}
\label{tab:cells}
\setlength{\fboxsep}{1.5pt}%
\begin{tabular}{llcccccccc}
\toprule
 & & \multicolumn{2}{c}{$F = 1$} & \multicolumn{2}{c}{$F = 2$} & \multicolumn{2}{c}{$F = 3$} & \multicolumn{2}{c}{$F = 4$} \\
\cmidrule(lr){3-4}\cmidrule(lr){5-6}\cmidrule(lr){7-8}\cmidrule(lr){9-10}
Size & $U$ & Transformer & Feedback & Transformer & Feedback & Transformer & Feedback & Transformer & Feedback \\
\midrule
60M & 12.5$N$ & \colorbox{black!9}{3.0141} & 3.0508 & \colorbox{black!9}{2.9060} & 2.9183 & \colorbox{black!9}{2.8631} & 2.8647 & 2.8404 & \colorbox[HTML]{D9D0E8}{2.8343} \\
 & 50$N$ & \colorbox{black!9}{2.8110} & 2.8199 & 2.7456 & \colorbox[HTML]{D9D0E8}{2.7424} & 2.7166 & \colorbox[HTML]{D9D0E8}{2.7074} & 2.6996 & \colorbox[HTML]{D9D0E8}{2.6866} \\
 & 100$N$ & 2.7428 & \colorbox[HTML]{D9D0E8}{2.7387} & 2.6918 & \colorbox[HTML]{D9D0E8}{2.6803} & 2.6689 & \colorbox[HTML]{D9D0E8}{2.6536} & 2.6553 & \colorbox[HTML]{D9D0E8}{2.6376} \\
 & 400$N$ & 2.6507 & \colorbox[HTML]{D9D0E8}{2.6343} & 2.6199 & \colorbox[HTML]{D9D0E8}{2.5987} & 2.6059 & \colorbox[HTML]{D9D0E8}{2.5826} & 2.5973 & \colorbox[HTML]{D9D0E8}{2.5737} \\
\midrule
134M & 12.5$N$ & \colorbox{black!9}{2.6927} & 2.7202 & \colorbox{black!9}{2.6073} & 2.6178 & \colorbox{black!9}{2.5697} & 2.5718 & 2.5483 & \colorbox[HTML]{D9D0E8}{2.5440} \\
 & 25$N$ & \colorbox{black!9}{2.5999} & 2.6126 & \colorbox{black!9}{2.5320} & 2.5333 & 2.5014 & \colorbox[HTML]{D9D0E8}{2.4964} & 2.4826 & \colorbox[HTML]{D9D0E8}{2.4743} \\
 & 50$N$ & \colorbox{black!9}{2.5281} & 2.5341 & 2.4716 & \colorbox[HTML]{D9D0E8}{2.4656} & 2.4452 & \colorbox[HTML]{D9D0E8}{2.4343} & 2.4292 & \colorbox[HTML]{D9D0E8}{2.4159} \\
 & 100$N$ & 2.4685 & \colorbox[HTML]{D9D0E8}{2.4670} & 2.4238 & \colorbox[HTML]{D9D0E8}{2.4155} & 2.4030 & \colorbox[HTML]{D9D0E8}{2.3905} & 2.3904 & \colorbox[HTML]{D9D0E8}{2.3750} \\
 & 200$N$ & 2.4234 & \colorbox[HTML]{D9D0E8}{2.4132} & 2.3880 & \colorbox[HTML]{D9D0E8}{2.3722} & 2.3711 & \colorbox[HTML]{D9D0E8}{2.3534} & 2.3610 & \colorbox[HTML]{D9D0E8}{2.3421} \\
 & 400$N$ & 2.3890 & \colorbox[HTML]{D9D0E8}{2.3741} & 2.3610 & \colorbox[HTML]{D9D0E8}{2.3428} & 2.3478 & \colorbox[HTML]{D9D0E8}{2.3281} & 2.3397 & \colorbox[HTML]{D9D0E8}{2.3190} \\
\midrule
271M & 12.5$N$ & \colorbox{black!9}{2.4691} & 2.4907 & \colorbox{black!9}{2.3957} & 2.4045 & \colorbox{black!9}{2.3629} & 2.3642 & 2.3428 & \colorbox[HTML]{D9D0E8}{2.3394} \\
 & 25$N$ & \colorbox{black!9}{2.3893} & 2.3993 & \colorbox{black!9}{2.3292} & 2.3308 & 2.3015 & \colorbox[HTML]{D9D0E8}{2.2974} & 2.2842 & \colorbox[HTML]{D9D0E8}{2.2777} \\
 & 50$N$ & \colorbox{black!9}{2.3283} & 2.3323 & 2.2774 & \colorbox[HTML]{D9D0E8}{2.2739} & 2.2540 & \colorbox[HTML]{D9D0E8}{2.2462} & 2.2397 & \colorbox[HTML]{D9D0E8}{2.2288} \\
 & 100$N$ & 2.2769 & \colorbox[HTML]{D9D0E8}{2.2744} & 2.2363 & \colorbox[HTML]{D9D0E8}{2.2283} & 2.2169 & \colorbox[HTML]{D9D0E8}{2.2058} & 2.2048 & \colorbox[HTML]{D9D0E8}{2.1921} \\
 & 200$N$ & 2.2374 & \colorbox[HTML]{D9D0E8}{2.2291} & 2.2042 & \colorbox[HTML]{D9D0E8}{2.1916} & 2.1885 & \colorbox[HTML]{D9D0E8}{2.1738} & 2.1786 & \colorbox[HTML]{D9D0E8}{2.1626} \\
 & 350$N$ & 2.2114 & \colorbox[HTML]{D9D0E8}{2.1988} & 2.1841 & \colorbox[HTML]{D9D0E8}{2.1681} & 2.1712 & \colorbox[HTML]{D9D0E8}{2.1533} & 2.1633 & \colorbox[HTML]{D9D0E8}{2.1444} \\
\midrule
778M & 12.5$N$ & \colorbox{black!9}{2.2054} & 2.2211 & \colorbox{black!9}{2.1434} & 2.1501 & \colorbox{black!9}{2.1136} & 2.1162 & 2.0950 & 2.0950 \\
 & 50$N$ & \colorbox{black!9}{2.0878} & 2.0893 & 2.0452 & \colorbox[HTML]{D9D0E8}{2.0421} & --- & --- & --- & --- \\
 & 100$N$ & 2.0448 & \colorbox[HTML]{D9D0E8}{2.0413} & --- & --- & --- & --- & --- & --- \\
\bottomrule
\end{tabular}%

\end{table}
\begin{table}[!ht]
\centering\footnotesize\setlength{\tabcolsep}{4pt}
\caption{\textbf{Measured multipliers.}
Top: compute multiplier $\kappa$ $\uparrow$ at equal size and unique data.
Bottom: data multiplier $\mu$ $\uparrow$ at equal size and epochs.
---: no such cell, or the read leaves the measured range (one to four Transformer epochs for $\kappa$, the largest $U$ for $\mu$).}
\label{tab:kappa}\label{tab:mu}
\begin{tabular}[t]{rcccccccccccc}
\toprule
 & \multicolumn{4}{c}{$F = 1$} & \multicolumn{4}{c}{$F = 2$} & \multicolumn{4}{c}{$F = 3$} \\
\cmidrule(lr){2-5}\cmidrule(lr){6-9}\cmidrule(lr){10-13}
UTPP & 60M & 134M & 271M & 778M & 60M & 134M & 271M & 778M & 60M & 134M & 271M & 778M \\
\midrule
12.5 & --- & --- & --- & --- & 0.92 & 0.92 & 0.92 & 0.93 & 0.98 & 0.98 & 0.98 & 0.97 \\
25 & --- & --- & --- & --- & --- & 0.99 & 0.98 & --- & --- & 1.08 & 1.07 & --- \\
50 & --- & --- & --- & --- & 1.05 & 1.10 & 1.06 & --- & 1.17 & 1.22 & 1.17 & --- \\
100 & 1.06 & 1.02 & 1.04 & --- & 1.23 & 1.18 & 1.18 & --- & --- & 1.33 & 1.30 & --- \\
200 & --- & 1.22 & 1.19 & --- & --- & 1.46 & 1.38 & --- & --- & --- & --- & --- \\
350 & --- & --- & 1.38 & --- & --- & --- & 1.68 & --- & --- & --- & --- & --- \\
400 & 1.45 & 1.45 & --- & --- & 1.91 & 1.79 & --- & --- & --- & --- & --- & --- \\
\bottomrule
\end{tabular}%
\par\vspace{6pt}
\begin{tabular}[t]{rcccccccccc}
\toprule
 & \multicolumn{4}{c}{1 epoch} & \multicolumn{3}{c}{2 epochs} & \multicolumn{3}{c}{3 epochs} \\
\cmidrule(lr){2-5}\cmidrule(lr){6-8}\cmidrule(lr){9-11}
$U$ & 60M & 134M & 271M & 778M & 60M & 134M & 271M & 60M & 134M & 271M \\
\midrule
12.5$N$ & 0.99 & 1.02 & 1.05 & 1.09 & 1.08 & 1.12 & 1.14 & 1.14 & 1.18 & 1.20 \\
25$N$ & --- & 1.10 & 1.14 & --- & --- & 1.19 & 1.23 & --- & 1.26 & 1.30 \\
50$N$ & 1.11 & 1.16 & 1.19 & 1.25 & 1.25 & 1.35 & 1.34 & 1.35 & 1.47 & 1.46 \\
100$N$ & 1.27 & 1.29 & 1.34 & --- & 1.46 & 1.49 & 1.53 & 1.60 & 1.67 & 1.69 \\
200$N$ & --- & 1.54 & 1.54 & --- & --- & 1.88 & --- & --- & --- & --- \\
\bottomrule
\end{tabular}%

\end{table}
\FloatBarrier
\clearpage

\section{The fitted law}\label{app:fit}
\begin{figure}[!t]
\centering
\includegraphics[width=\linewidth]{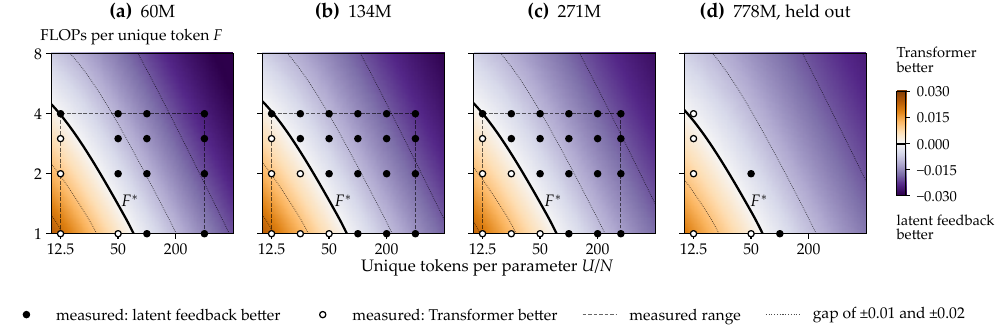}
\vspace{-8pt}
\caption{\textbf{The loss gap at equal compute across unique data and budget.}
Markers are measured cells; everywhere else, the map is extrapolated with the law of Section~\ref{sec:law}.
Latent feedback wins above the solid line, the crossover $F^\ast$, which falls as unique data grows and hardly moves with size.}
\label{fig:decision}
\end{figure}
Equation~\ref{eq:law-colored} is Equation~\ref{eq:law} in its best form (Table~\ref{tab:forms}).
It is the coupled form, with latent feedback's capacity growing with TPP and with size.
It is fitted jointly to both models' 128 cells at 60M, 134M and 271M.
Each symbol has one of four kinds (Table~\ref{tab:symbols}).
\In{\textbf{Inputs}} are chosen by whoever uses the fitted law.
\Fit{\textbf{Fitted coefficients}} are estimated once, jointly, from the 128 cells.
\Fix{\textbf{Fixed constants}} are measured or set by design and never fitted.
Black symbols are \textbf{derived}: computed from the other three kinds.
\begin{gather}
L_{\In{s}}(\In{N}, \In{U}, \In{F}) = \Fit{E} + \Big[\Big(\frac{\Fit{A}}{(v_{\In{s}}\,n)^{\Fit{\alpha}}}\Big)^{\Fit{q}} + \Big(\frac{\Fit{B}}{d_{\In{s}}^{\,\Fit{\beta}}}\Big)^{\Fit{q}}\Big]^{1/\Fit{q}}, \qquad
v_{\In{s}} = \Fit{v_0}\Big(\frac{d_{\In{s}}}{n}\Big)^{\Fit{\omega}} n^{\Fit{\gamma}}, \label{eq:law-colored}\\
D_{\In{s}} = \In{U} \Big[1 + R^\ast\big(1 - e^{-(R-1)/R^\ast}\big)\Big], \qquad
R = \frac{\In{F}}{\Fix{c_s}}, \qquad R^\ast = \Fit{R^\ast_T}\Big(\frac{u}{n}\Big)^{\Fit{\rho}}, \notag\\
n = \frac{\In{N}}{\Fix{N_0}}, \qquad u = \frac{\In{U}}{\Fix{D_0}}, \qquad d_{\In{s}} = \frac{D_{\In{s}}}{\Fix{D_0}}, \qquad
\Fix{N_0 = 10^{8}}, \qquad \Fix{D_0 = 100\,N_0 = 10^{10}}. \notag
\end{gather}
Below one epoch the bracket is replaced by $R$.
The Transformer $T$ has $\Fix{v_T = 1}$.
Every size is measured against one reference run: $N_0 = 10^8$ non-embedding parameters trained on $D_0 = 10^{10}$ effective tokens, that is, 100 tokens per parameter.
So $d_s/n$ counts effective tokens per parameter in units of 100, and $A$, $B$, $v_0$ and $R^\ast_T$ are the values of their terms at the reference run.
The reference is a choice of units: moving it rescales these four coefficients and changes no prediction.

\begin{table}[!ht]
\centering\footnotesize\setlength{\tabcolsep}{6pt}
\caption{\textbf{The fitted law beyond the grid:} predicted losses along one curve of Figure~\ref{fig:extrap} (a) to (c), with $\kappa$.
Each block extends one axis from a grid cell (its first row); sizes above 778M count non-embedding parameters.
$\infty$: no number of epochs lets the Transformer match.}
\label{tab:extrap}
\begin{tabular}{llrrccc}
\toprule
Axis extended & Size & $U$ & $F$ & Transformer $\downarrow$ & Feedback $\downarrow$ & $\kappa$ $\uparrow$ \\
\midrule
Larger model & 271M & 100$N$ & 2 & 2.2356 & 2.2266 & 1.22 \\
 & 778M & 100$N$ & 2 & 2.0116 & 2.0025 & 1.27 \\
 & 3B & 100$N$ & 2 & 1.7735 & 1.7639 & 1.39 \\
 & 10B & 100$N$ & 2 & 1.6085 & 1.5988 & 1.51 \\
 & 100B & 100$N$ & 2 & 1.3505 & 1.3410 & 1.83 \\
\midrule
More unique data & 271M & 350$N$ & 1 & 2.2095 & 2.1967 & 1.41 \\
 & 271M & 1,000$N$ & 1 & 2.1754 & 2.1561 & 2.50 \\
 & 271M & 4,000$N$ & 1 & 2.1485 & 2.1227 & $>$10 \\
 & 271M & 16,000$N$ & 1 & 2.1341 & 2.1035 & $\infty$ \\
\midrule
More repeats & 271M & 50$N$ & 4 & 2.2399 & 2.2302 & 1.27 \\
 & 271M & 50$N$ & 8 & 2.2150 & 2.1999 & 1.97 \\
 & 271M & 50$N$ & 16 & 2.1997 & 2.1803 & $\infty$ \\
 & 271M & 50$N$ & 32 & 2.1926 & 2.1700 & $\infty$ \\
\bottomrule
\end{tabular}%

\end{table}

\begin{table}[!ht]
\centering\footnotesize\setlength{\tabcolsep}{4pt}
\caption{\textbf{Forms of Equation~\ref{eq:law}} on the cells at 60M, 134M and 271M, with $E \ge 0.5$.
RMSE over the cells and over the 64 equal-FLOPs gaps, in sample and leave-one-out (LOO).
$v(\mathrm{TPP})$: $v$ grows as a power of TPP; $v(\mathrm{TPP}, N)$: also of $N$. Skaling equals the coupled form ($k = 1/q$); Kaplan ties $q$ to $1/\beta$.}
\label{tab:forms}
\begin{tabular}{llcccccc}
\toprule
Form & Fit & Cells & Coefficients & RMSE $\downarrow$ & LOO $\downarrow$ & Gap RMS $\downarrow$ & Gap LOO $\downarrow$ \\
\midrule
\multicolumn{8}{l}{\textit{Transformer only}} \\
Additive & two-phase & 64 & 7 & 0.0086 & 0.0103 & --- & --- \\
Additive & joint & 64 & 7 & 0.0070 & 0.0080 & --- & --- \\
Coupled & two-phase & 64 & 8 & 0.0034 & 0.0037 & --- & --- \\
Coupled & joint & 64 & 8 & \textbf{0.0019} & \textbf{0.0024} & --- & --- \\
\midrule
\multicolumn{8}{l}{\textit{Both arms}} \\
Additive & joint & 128 & 8 & 0.0079 & 0.0087 & 0.0068 & 0.0082 \\
Additive, $v(\mathrm{TPP})$ & joint & 128 & 9 & 0.0076 & 0.0084 & 0.0050 & 0.0060 \\
Coupled & joint & 128 & 9 & 0.0027 & 0.0032 & 0.0035 & 0.0044 \\
Coupled, $v(\mathrm{TPP})$ & joint & 128 & 10 & \textbf{0.0025} & \textbf{0.0030} & 0.0031 & 0.0039 \\
Coupled, $v(\mathrm{TPP}, N)$ & joint & 128 & 11 & \textbf{0.0025} & \textbf{0.0030} & \textbf{0.0029} & \textbf{0.0037} \\
Skaling, $v(\mathrm{TPP}, N)$ & joint & 128 & 11 & \textbf{0.0025} & \textbf{0.0030} & \textbf{0.0029} & \textbf{0.0037} \\
Kaplan, $v(\mathrm{TPP}, N)$ & joint & 128 & 10 & 0.0149 & 0.0161 & 0.0139 & 0.0148 \\
\bottomrule
\end{tabular}%

\end{table}

\begin{itemize}
\item \textbf{One fit per model family.}
$E, A, \alpha, B, \beta$ and $q$ belong to one shape ladder, dataset, tokenizer, optimizer and tuning recipe.
Every schedule shares them, and a new family refits them.
The batch is not fixed: at small $U$ it follows the optimum of \citet{bergsma2025powerlines}.
Its $\beta_2$ then moves to keep its half-life in tokens~\citep{marek2025small} (Appendix~\ref{app:config}).
\item \textbf{The count of $N$.}
We count non-embedding parameters in both $U/N$ and the size term~\citep{kaplan2020scaling,li2025steplaw,li2025farseer,kimi2025linear}.
\citet{pearce2024reconciling} trace part of the disagreement between Kaplan and Chinchilla to this count.
The count matters at small scale: at 60M, embeddings are 55\% of the parameters.
Counting all parameters, or adding the output head, fits the Transformer cells about as well.
The joint coupled RMSE is then 0.0014 against 0.0018, but $\alpha$ moves to 0.24 or 0.18.
\item \textbf{The size term.}
Fitting $\gamma$ barely changes the in-sample error, and an F-test of $\gamma$ against $\gamma = 0$ gives $p = 0.054$.
Without $\gamma$, however, the law underrates latent feedback in all seven held-out 778M rows, and its gap error grows 2.5 times.
These rows come from three main runs, so they give only three independent signs, not seven.
We are training larger models to test the size term further.
\item \textbf{The loss floor.}
Three sizes do not pin it: $E$ sits at its lower bound of 0.5 in every form.
Freed from its bound, $E$ falls to 0 to 0.3 at equal error.
The exponent $\alpha$ then moves from 0.11 to 0.07 to 0.10.
\end{itemize}

\section{Derivation of the crossover}\label{app:crossover}
This appendix derives Equation~\ref{eq:gap}, gives the exact condition for the crossover budget $F^\ast$, and shows that the law has at most one crossover.

\textbf{Setup.}
Fix $N$, $U$ and $F$.
The Transformer has $v = 1$ and trains $R = F$ epochs; its terms are $X = A/N^{\alpha}$ and $Y = B/D(F)^{\beta}$, and its reducible loss is $S = (X^q + Y^q)^{1/q}$.
A schedule $s$ multiplies each term by a factor:
\begin{equation}
X_s = X\,v_s^{-\alpha}, \qquad Y_s = Y\,\Big(\frac{D(F)}{D(F/c_s)}\Big)^{\beta} = Y\,c_s^{\,\bar\varepsilon\beta}, \qquad
\bar\varepsilon = \frac{1}{\ln c_s}\int_{F/c_s}^{F} \varepsilon(R)\,\mathrm{d}\ln R. \label{eq:factors}
\end{equation}
Here $\varepsilon(R) = \mathrm{d}\ln D/\mathrm{d}\ln R$ is the elasticity of effective data to epochs.
So the $\bar\varepsilon$ of Equation~\ref{eq:gap}, with $D_T = D(F)$ and $D_s = D(F/c_s)$, is the mean of this elasticity over the epochs that the schedule gives up.

\textbf{First order.}
Because $\ln S = \frac{1}{q}\ln(X^q + Y^q)$, its derivatives are $\partial\ln S/\partial\ln X = w$ and $\partial\ln S/\partial\ln Y = 1 - w$, with $w = X^q/(X^q + Y^q)$.
So, to first order in $\ln v_s$ and $\ln c_s$,
\begin{equation}
\ln\frac{S_s}{S} \approx -\,w\,\alpha\ln v_s \;+\; (1-w)\,\bar\varepsilon\,\beta\ln c_s. \label{eq:lngap}
\end{equation}
Equation~\ref{eq:gap} follows, since $L_s - L_T = S_s - S \approx S\ln(S_s/S)$.

\textbf{Elasticity.}
From Equation~\ref{eq:law}, with $x = (R-1)/R^\ast$,
\begin{equation}
\varepsilon(R) = \frac{R\,e^{-x}}{1 + R^\ast(1 - e^{-x})} \ \text{ for } R \geq 1, \qquad \varepsilon(R) = 1 \ \text{ for } R \leq 1. \label{eq:eps}
\end{equation}
It is 1 at one epoch and falls to 0 as repeats saturate.
It never rises on the way: $\mathrm{d}\ln\varepsilon/\mathrm{d}\ln R = 1 - R/R^\ast - \varepsilon$, and $\varepsilon \geq e^{-x} \geq 1 - x > 1 - R/R^\ast$, because the denominator is at most $1 + R^\ast x = R$.

\textbf{Exact crossover.}
The two losses are equal when $X_s^q + Y_s^q = X^q + Y^q$.
With Equation~\ref{eq:factors}, this is
\begin{equation}
\Big(\frac{X}{Y}\Big)^{q} = \frac{c_s^{\,q\bar\varepsilon\beta} - 1}{1 - v_s^{-q\alpha}}. \label{eq:crossover}
\end{equation}
No approximation enters.
To first order, its right side is $\bar\varepsilon\,\beta\ln c_s / (\alpha\ln v_s)$: data cost over capacity gain.

\textbf{One crossover.}
Hold $N$ and $U$ and raise $F$.
The left side of Equation~\ref{eq:crossover} rises, because $D(F)$ grows, so $Y$ falls while $X$ stays.
The right side falls: $\bar\varepsilon$ falls because $\varepsilon$ does, and $v_s$ does not shrink, since $\omega > 0$.
Therefore the two sides meet at most once, and the schedule loses below $F^\ast$ and wins above it.
Moreover, the right side tends to 0 as repeats saturate while the left side stays positive.
So under the law, a costlier schedule with $v_s > 1$ wins at a large enough budget.

\textbf{Data per parameter.}
More unique data lowers $Y$ at every $F$, which raises the left side and moves $F^\ast$ earlier.
However, it also raises $R^\ast$, which slows saturation and moves $F^\ast$ later.
In the fitted law the first effect wins: at 134M, $F^\ast$ falls from 4.1 at $U = 12.5N$ to 1.5 at $50N$ (Figure~\ref{fig:decision}).

\textbf{Accuracy.}
Equation~\ref{eq:crossover} reproduces every crossover of Figure~\ref{fig:decision}.
Equation~\ref{eq:gap} stays within 0.002 of the fitted law's gap for $U = 12.5N$ to $100N$ and $F = 1$ to 4, always in favour of latent feedback.
So its crossover comes 7\% to 12\% early: 3.8 against 4.1 at 134M and $U = 12.5N$.
The bias arises because the first-order ratio lies below the exact right side of Equation~\ref{eq:crossover}.

\section{Model and training configuration}\label{app:config}
All sizes share one recipe (Table~\ref{tab:config}).
They differ only in shape, learning rate, weight decay, schedule lengths and, at small $U$, the batch.

\textbf{Learning rate.}
The learning rate $\eta$ is that of the matrix optimizer.
The embedding and the output head train at $0.905\,\eta$ and $0.08\,\eta$.

\textbf{Weight decay.}
It follows a timescale rule $\tau = r\,a\,\mathrm{TPP}^{b}$ in training tokens per parameter (TPP).
The rule has $(a, b) = (0.403, -0.483)$ for the Transformer and $(0.751, -0.577)$ for latent feedback.
We re-centre $r$ per arm on the 134M cells.

\textbf{Tuning grid.}
Each model gets three learning rates and two or three $\tau$ levels per cell (Table~\ref{tab:config}).
We add the next point wherever the best one sits on an edge of the grid.
A cell's value is the minimum of a quadratic fitted to its points if that minimum lies inside the grid, and otherwise the best point.

\textbf{Decay branches.}
Each branch keeps its main run's $\eta$ and $\lambda$, as in other warmup-stable-decay scaling studies~\citep{hagele2024scaling}.
On an earlier pool we checked this against a separate ladder, whose runs were tuned on their own grids for about two epochs.
In seven cells at two sizes and 1 to 2.3 epochs, five of the Transformer and two of latent feedback, the branches match those runs within 0.0012, with a mean difference of 0.0003.

\textbf{Batch.}
The batch keeps each cell near the optimum of \citet{bergsma2025powerlines}.
It stays between 0.46 and 1.72 times that optimum at every size.
At 134M and 271M it is 0.25M tokens up to $U = 25N$ and 0.5M from $50N$; 778M uses 0.5M (Table~\ref{tab:config}).
Both models share each cell's batch; one 60M cell run at both batches moves the gap by 0.002.
We set $\beta_2$ by its half-life in tokens~\citep{marek2025small}.

\begin{table}[!ht]
\centering\footnotesize\setlength{\tabcolsep}{5pt}
\caption{Configuration of the runs; the latent feedback Transformer adds the feedback map $\F{W}$ (Equation~\ref{eq:fbt}).
Learning rate: centre of the grid, Transformer / latent feedback, one step of $\sqrt{2}$ higher at $U = 12.5N$ except at 134M.
Warmup doubles with each halving of the batch.}
\label{tab:config}
\begin{tabular}{lcccc}
\toprule
 & 60M & 134M & 271M & 778M \\
\midrule
Parameters & 60,041,728 & 134,108,928 & 271,094,784 & 777,862,656 \\
Non-embedding parameters & 27,273,728 & 84,956,928 & 205,558,784 & 679,558,656 \\
Layers & 8 & 12 & 16 & 24 \\
Width & 512 & 768 & 1024 & 1536 \\
Attention heads & 4 & 6 & 8 & 12 \\
MLP hidden width & 1536 & 2048 & 2816 & 4096 \\
\midrule
$U/N$ & 12.5, 50, 100, 400 & 12.5 to 400 & 12.5 to 350 & 12.5, 50, 100 \\
Main run length $F$, by $U/N$ & 4 & 4 & 4 & 4, 2, 1 \\
Tokens per update$^a$ & 0.125M / 0.25M & 0.25M / 0.5M & 0.25M / 0.5M & 0.5M \\
$\beta_2$, AdamW and NorMuon$^a$ & 0.987 / 0.975 & 0.975 / 0.95 & 0.975 / 0.95 & 0.95 \\
Warmup updates at 0.5M & 115 & 256 & 517 & 1,484 \\
\midrule
Latent feedback: one-pass probability $p$ & \multicolumn{4}{c}{0.15} \\
Latent feedback: cost $c$ & 1.1970 & 1.2317 & 1.2514 & 1.2677 \\
\midrule
Learning rate $\eta$ & 0.0252 & 0.0163 & 0.0111 / 0.0157 & 0.0043 / 0.0061 \\
Tuning grid in $\eta$$^b$ & \multicolumn{4}{c}{$\eta \times \sqrt{2}^{\pm 1}$} \\
Tuning grid in $\tau$$^b$ & $\tau \times 2^{\pm 1}$ & $\tau \times 2^{\pm 1}$ & $\tau \times 2^{-1}, 2^{-2}$ & --- \\
\midrule
Data & \multicolumn{4}{c}{Stratified Dolma~3 (Appendix~\ref{app:pools})} \\
Tokenizer & \multicolumn{4}{c}{Llama 2 (32,000 tokens)} \\
Sequence length & \multicolumn{4}{c}{2048 tokens} \\
Epoch order & \multicolumn{4}{c}{One fixed order, the same in every epoch} \\
Block & \multicolumn{4}{c}{Pre-norm RMSNorm, rotary positions, QK-norm, SwiGLU; head width 128} \\
Embeddings & \multicolumn{4}{c}{Untied input and output, no weight decay} \\
Initialization & \multicolumn{4}{c}{OLMo 2~\citep{olmo2025olmo2}} \\
Matrix optimizer & \multicolumn{4}{c}{NorMuon, momentum 0.95, from 0.85 over the first 300 updates at 0.5M} \\
Embedding and head & \multicolumn{4}{c}{AdamW, $\beta_1 = 0.9$} \\
Schedule & \multicolumn{4}{c}{Warmup-stable-decay; linear decay to zero over the last 20\% of each budget} \\
Other & \multicolumn{4}{c}{Gradient clipping at 1.0; bfloat16} \\
\bottomrule
\end{tabular}
\\[2pt]{\footnotesize $^a$At low $U$ / above: low is $U = 12.5N$ at 60M, and $12.5N$ and $25N$ at 134M and 271M. $^b$Plus the next point out wherever a cell's best point sits on an edge; $\tau$ points at the centre $\eta$, and at 271M only at $U = 12.5N$ and $50N$.}
\end{table}

\textbf{Choice of $p$.}
On earlier pools, $p$ from 0.01 to 0.31 changed the loss at equal compute by under 0.003.
With $p = 0$, the passes fell into a two-cycle, so decoding failed.

\clearpage
\section{Downstream suites}\label{app:checks}
\textbf{Weighting the downstream suites.}
Table~\ref{tab:main} lists the downstream BPB of every cell.
A plain mean over the 37 suites is dominated by yes/no and arithmetic suites, whose seed noise reaches 0.24 BPB.
So we weight each suite by the inverse of its seed variance.
Let $b_{r,j}$ be run $r$'s BPB on suite $j$, the mean over the suite's tasks.
The seed variance $\sigma_j^2$ is the variance of $b_{r,j}$ across seeds 0 to 2, averaged over 16 groups of runs at 134M (both models, $U = 12.5N$ and $100N$, $F = 1$ to 4).
A run's downstream score is then
\begin{equation}
b_r = \frac{\sum_j w_j\, b_{r,j}}{\sum_j w_j}, \qquad w_j = \frac{1}{\sigma_j^2}.
\end{equation}

\section{Nested pools}\label{app:pools}

\textbf{Construction.}
Each raw file of the Dolma~3 sample holds one of 224 source-topic buckets.
We keep the first copy of each exact duplicate, a document with the same token sequence.
We drop documents that share a paragraph of 13+ words with Paloma's validation or test sets~\citep{magnusson2023paloma}.
These are 0.16\% of the documents.
We shuffle documents within each bucket.
We then merge the buckets so that each advances along the stream at the rate of its token share.
Every prefix is then a stratified sample of the stream, so pools differ only in size.
From 1B tokens on, a prefix's bucket mixture lies within 0.003 of the whole stream's in total variation.

\end{document}